%% file: neurips_2026.tex
\documentclass{article}

\PassOptionsToPackage{square, authoryear}{natbib}

 \usepackage[main, final]{neurips_2026}

\usepackage[utf8]{inputenc} %
\usepackage[T1]{fontenc}    %
\usepackage{hyperref}       %
\usepackage{url}            %
\usepackage{booktabs}       %
\usepackage{amsfonts}       %
\usepackage{nicefrac}       %
\usepackage{microtype}      %
\usepackage{xcolor}         %

\usepackage{xspace}
\newcommand{\method}{\texttt{ProQuant}\xspace}
\title{\method: Progressive Quantization-aware Training for Edge MLLMs}

\author{%
  \textbf{Yufei Xue$^{1}$ \quad
  Yushi Huang$^{1}$ \quad
  Jiawei Shao$^{2,3}$%
  \thanks{Corresponding authors.} \quad
  Pingcheng Dong$^{1}$} \\
  \textbf{Yonghao Tan$^{1}$ \quad
  Shiyao Li$^{4}$ \quad
  Kwang-Ting Cheng$^{1}$ \quad
  Jun Zhang$^{1}$\footnotemark[1]} \\
  $^{1}$The Hong Kong University of Science and Technology \\
  $^{2}$Northwestern Polytechnical University \\
  $^{3}$Institute of Artificial Intelligence (TeleAI), China Telecom \\
  $^{4}$Tsinghua University \\
  \texttt{jiawei.shao@nwpu.edu.cn},
  \texttt{eejzhang@ust.hk}
}

\input{utils/packages}

\begin{document}

\maketitle

\begin{abstract}
    Multi-modal large language models (MLLMs) with impressive perception and understanding capabilities will enable a range of downstream applications. However, their substantial computational and memory requirements hinder the real-world implementation, particularly for resource-constrained edge devices. Quantization has emerged as an effective solution for deploying large models. While post-training quantization (PTQ) successfully retains the performance at moderate bit-width (\textit{e.g.}, W8A8), it suffers from severe performance degradation under low-bit quantization (\textit{e.g.}, W4A8), especially for small-scale models that are more suitable for edge deployment. In this paper, we propose \texttt{ProQuant}, a novel weight-activation quantization-aware training (QAT) framework tailored for low-bit MLLMs. First, to align the limited finetuning dataset distribution with the pretrained dataset distribution, we propose a self-relabeling (SRL) scheme to resample the responses of the open-source multimodal dataset. Second, we are the \textit{first} to take a decoupled view of multi-modal (MM) understanding and language modeling abilities. We propose a progressive block activation (PBA) mechanism to prioritize the MM understanding recovery. We also introduce a vision-sensitive factor $\vec{\tau}$ that effectively identifies the MM-critical layers. Extensive experiments across various benchmarks, covering models with 2B$\sim$8B parameters, show that \texttt{ProQuant} significantly outperforms existing methods. For example, our W4A8 \texttt{Qwen3-VL-2B-Instruct} improves average accuracy by impressively 6.6\% and achieves performance comparable to full precision counterparts with only $\sim$1\% performance drop. Code is available at \url{https://github.com/yufeixue-ai/ProQuant/}.
\end{abstract}

\input{docs/introduction}
\input{docs/related_works}

\input{docs/method}
\input{docs/experiments}

\input{docs/conclusion}

\begin{ack}
We thank the anonymous reviewers for their constructive feedback. This work received no external funding or third-party support. The authors declare no competing interests.
\end{ack}

\bibliographystyle{unsrtnat}
\bibliography{main}

\clearpage
\input{docs/appendix}

\end{document}

%% file: utils/packages.tex
\usepackage{xcolor}
\usepackage{colortbl}
\definecolor{myyellow}{RGB}{250,225,165}
\definecolor{mygreen}{RGB}{166,235,209}
\definecolor{myblue}{RGB}{194,209,241}
\definecolor{myred}{RGB}{239,193,192}
\definecolor{mypurple}{RGB}{212,191,240}

\definecolor{citepcolor}{HTML}{0071BC}
\definecolor{linkcolor}{HTML}{ED1C24}
\definecolor{mydarkgreen}{rgb}{0.02,0.6,0.02}

\usepackage{array}
\usepackage{amssymb}
\usepackage{amsmath}
\usepackage{enumitem}
\usepackage{multirow}
\usepackage{booktabs}
\usepackage{pifont}
\usepackage{makecell}
\usepackage{graphicx}
\usepackage{wrapfig}  
\usepackage{algorithm}
\usepackage{algpseudocode}
\usepackage[font=small,labelfont=bf,tableposition=top]{caption}
\usepackage[font=scriptsize,skip=1pt,subrefformat=parens]{subcaption}
\usepackage{pifont}  %

\usepackage{tcolorbox}
\tcbuselibrary{breakable, skins} 
\newtcolorbox{tocbox}{
  enhanced,
  breakable,
  colback=white,
  colframe=black,
  boxrule=0.4pt, 
  left=8pt,right=8pt,top=8pt,bottom=8pt,
}
\usepackage{lipsum}  %
\usepackage{pgfplots} %
\pgfplotsset{compat=1.18} %
\usetikzlibrary{shadows}
\usetikzlibrary{shadows.blur} %

\newcommand{\bg}{\cellcolor{myblue!25}} %
\newcommand{\rowbg}{\rowcolor{myblue!25}} %

\newcommand{\speedupbg}{\cellcolor{myblue!25}} %
\newcommand{\savingbg}{\cellcolor{myred!25}} %

\newcommand{\eg}{e.g.\@}
\newcommand{\ie}{i.e.\@}

%% file: docs/introduction.tex
\section{Introduction}\label{sec:introduction}

Extending large language models (LLMs) beyond text to additional modalities~\citep{bai2025qwen3vl,li2024llava-ov,xiaomi2025mimovl,zhu2025internvl3,wu2024deepseekvl2,team2026ovis,li2026pace} has achieved remarkable successes and enabled a wide range of applications, including chatbots~\citep{bai2025qwen3vl}, autonomous driving~\citep{tian2024drivevlm}, and AR/VR glasses~\citep{chen2024videollm-online}. However, the large memory footprint and substantial computational overhead~\citep{huang2026modesacceleratingmixtureofexpertsmultimodal,lv2025llmcbenchmarkingvisionlanguagemodel} of multi-modal LLMs (MLLMs) present major obstacles to deployment in resource-constrained environments, especially on edge devices \citep{an2026aiflow,shao2025aiflow,yuan2025task}. Quantization~\citep{gholami2022quant-survey} alleviates this issue by reducing the precision of both weights and activations, thereby accelerating memory access and computation. Post-training quantization (PTQ)~\citep{xiao2023smoothquant,ashkboos2024quarot,liu2024spinquant,huang2024tfmq-dm,huang2025temporal-feature-matters,gong2024llmc,gong2025pushing} typically delivers near-lossless performance at moderate precision (\eg, W8A8), yet its accuracy degrades markedly when pushed to lower-bit settings (\eg, W4A4/W4A8). Additionally, small-scale models (2B/4B), which are more practical for edge deployment, typically suffer more pronounced performance degradation after quantization~\citep{chen2025qat-scaling-law}.

In contrast, quantization-aware training (QAT) that incorporates quantization during training has emerged as a promising solution to recover the model performance under low-bit quantization settings~\citep{liu2023llm,du2024bitdistiller,chen2025qat-scaling-law,huang2025qvgen}. The QAT solutions for LLM primarily target the training data corpus~\citep{liu2023llm}, training efficiency~\citep{chen2025efficientqat}, or distillation paradigm~\citep{du2024bitdistiller}. In the context of MLLM QAT, additional challenges arise. First, preserving the zero-shot generalization of compressed models hinges on high-quality fine-tuning data that are sufficiently broad and well-aligned with the original pre-training distribution. However, open-source datasets are often domain-narrow or distribution-shifted~\citep{liu2023llm}, and multi-modal (MM) data are far more difficult to collect than text-only data~\citep{guo2025speedq}. Second, quantization degrades both MM understanding and language modeling capabilities. A natural question is whether these two capabilities should be recovered jointly, or whether one should be prioritized during training. 

\begin{wrapfigure}{r}{0.45\linewidth}
    \centering
    \vspace{-2.5em}
    \includegraphics[width=\linewidth]{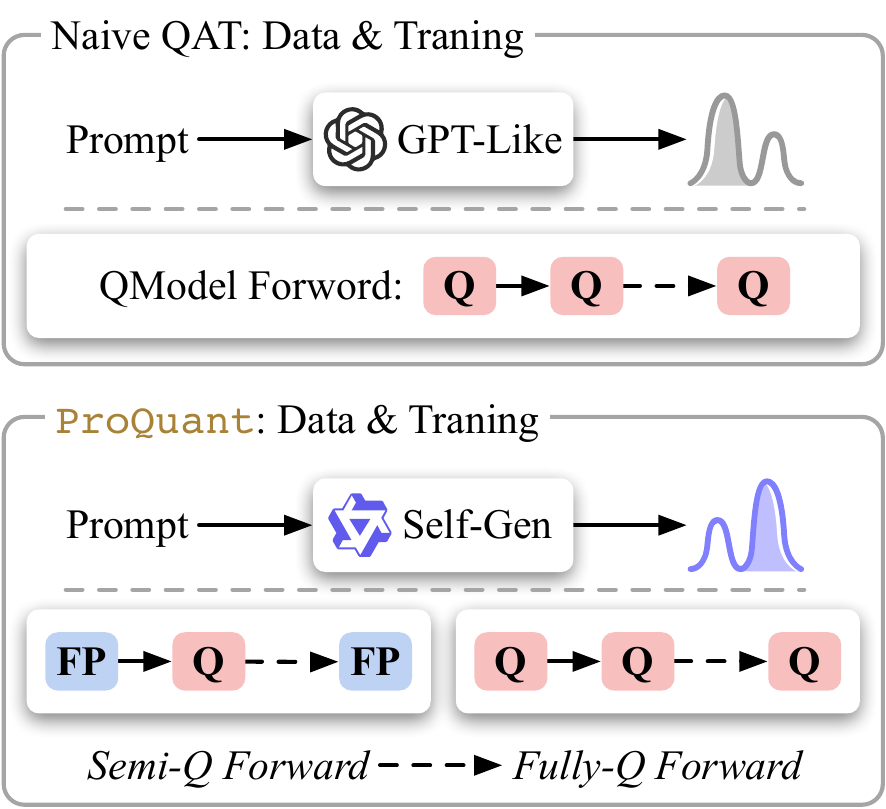}
    \caption{Naive QAT \textit{v.s.} \method. }
    \vspace{-2.5em}
    \label{fig:teaser}
\end{wrapfigure}

Considering the challenges discussed above, we propose \method, a novel QAT framework tailored to small-scale MLLMs, as illustrated in Figure~\ref{fig:teaser}. First, we propose a self-relabeling (SRL) scheme that re-annotates the training set to better align the fine-tuning data distribution with the model distribution. Second, we introduce a progressive block activation (PBA) mechanism that prioritizes recovering the MM understanding ability by first enabling quantization for vision-sensitive layers while keeping the remaining blocks in full precision during the forward pass. As training progresses, we gradually activate quantization for increasingly text-sensitive layers, thereby restoring language modeling capability. The core contributions are summarized as follows:
\begin{itemize}[leftmargin=*, noitemsep, label=$\triangleright$]
    \item We propose \method, a dedicated QAT framework for MLLMs. To the best of our knowledge, it is \textit{the first} weight-activation QAT framework for MLLMs, enabling effective low-bit (\eg, W4A8/W4A4) quantization.
    \item To address the scarcity of high-quality MM data, we introduce the SRL scheme, which re-annotates the fine-tuning data to better match the model distribution, thereby reducing data mismatch and improving post-quantization generalization.
    \item We take a disentangled view of MM understanding and language modeling capabilities during quantization. Particularly, we propose the PBA mechanism to prioritize recovering MM understanding first, and then gradually restore language modeling ability, instead of jointly optimizing both from the outset.
    \item Extensive experiments show that \method achieves state-of-the-art (SOTA) performance across various models and quantization settings. Notably, \method attains near-lossless compression ($\sim$1\% performance drop) under W4A8 quantization and substantially outperforms baselines under W4A4, where PTQ methods often collapse.
\end{itemize}

%% file: docs/related_works.tex
\section{Related Works}\label{sec:related_works}

\subsection{Post-training quantization for (M)LLMs.} 
PTQ has become a widely adopted technique for compressing and accelerating large models. In the context of LLMs, SmoothQuant-style methods~\citep{xiao2023smoothquant,lin2023awq,shao2023omniquant} mitigate activation outliers in transformer blocks via smoothing-based transformations. More recently, incoherence-processing-based approaches~\citep{ashkboos2024quarot,liu2024spinquant,hu2025ostquant} reshape weight distributions into more quantization-friendly ones, albeit at the cost of introducing online activation rotations. An orthogonal line of work, namely Hessian-informed techniques~\citep{frantar2022gptq,li2025gptqv2}, performs iterative weight quantization with closed-form updates. However, PTQ for VLMs is considerably more challenging due to heterogeneous features and mismatched distributions between text and vision modalities. Recent efforts such as MBQ~\citep{li2025mbq} and VLMQ~\citep{xue2025vlmq} attribute performance degradation to modality discrepancy and apply asymmetric treatments to vision and text tokens during quantization. MQuant~\citep{yu2025mquant} further extends rotation-based quantization by introducing modality-specific static activation quantization and attention-invariant flexible switching to better handle outliers.

\subsection{Quantization-aware training for (M)LLMs.} 
QAT can further improve quantized models' accuracy beyond PTQ, particularly under low-bit settings. LLM-QAT~\citep{liu2023llm} pioneers the QAT paradigm for language transformers. Efficient-QAT~\citep{chen2025efficientqat} improves training efficiency via block-wise initialization of quantization parameters. BitDistiller~\citep{du2024bitdistiller} and SPEED-Q~\citep{guo2025speedq} investigate knowledge distillation within the quantization context. Other learning-based works~\citep{wang2024q-vlm,xie2024qslaw} enhance quantized performance by exploiting cross-layer dependencies and refining the data corpus, respectively. However, except for LLM-QAT, the above methods primarily target weight-only quantization. To the best of our knowledge, no previous work has systematically explored weight-activation quantization for MLLMs. See Appendix~\ref{app:background} for more details.

%% file: docs/method.tex
\section{\method}\label{sec:method}
\begin{figure}[t]
    \centering
    \includegraphics[width=\linewidth]{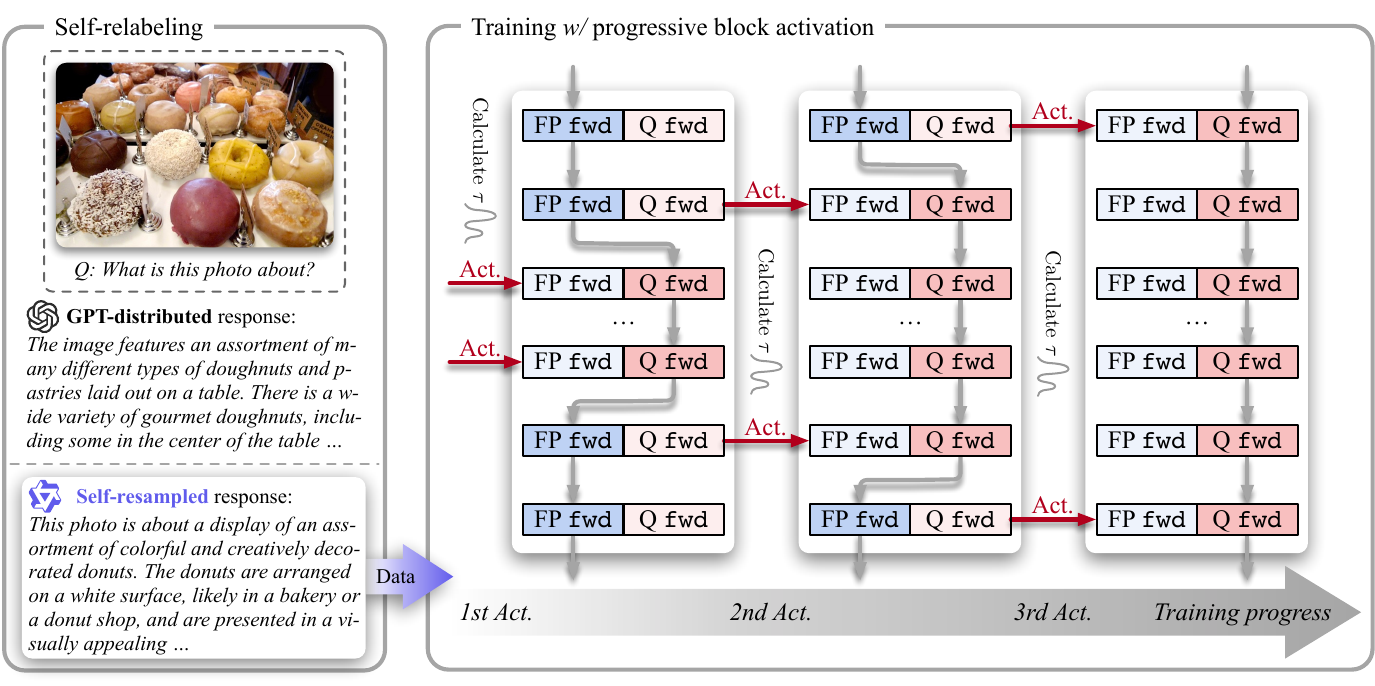}
    \caption{Overview of the proposed \method. Left: examples of responses from \texttt{GPT-4}~\citep{bubeck2023gpt4} and re-sampled via the SRL scheme (Section~\ref{sec:srl}) using \texttt{Qwen3-VL-2B-Instruct} (see more details in Appendix~\ref{app:method}). Right: QAT with the PBA mechanism, where quantization is first activated for decoding layers with the larger vision-sensitivity factor $\tau$. As training proceeds, quantization is gradually enabled for the remaining layers in descending order of $\tau$ (Section~\ref{sec:pba}).}
    \label{fig:overview}
\end{figure}

In this section, we present \method, a dedicated QAT framework for MLLMs that improves both the fine-tuning data distribution and the training pipeline. Specifically, 1) to obtain high-quality fine-tuning data that are closely aligned with the pre-training distribution, we discard the original labels in open-source MM datasets and resample responses from the model to quantize. Then, 2) We adopt a disentangled view of MM understanding and language modeling in MLLMs, and prioritize recovering MM understanding first via the proposed PBA mechanism. The vision-sensitive modules in the LLM backbone are identified using our proposed factor $\vec{\tau}$.

\subsection{Data Augmentation via Self-Relabeling}\label{sec:srl}
\textbf{SRL scheme.}
To align the distribution of the limited fine-tuning data with the pretrained data distribution, we resample responses for open-source MM datasets, which we refer to as the SRL scheme. Specifically, we re-generate responses with the target model using default sampling parameters until reaching either the \texttt{[EOS]} token or the maximum generation length. In Figure~\ref{fig:overview} (left), We present responses of \texttt{GPT-4}~\citep{bubeck2023gpt4} and \texttt{Qwen3-VL-2B-Instruct} triggered by prompts from the COCO caption dataset~\citep{liu2024llava1.5}. The two sets of responses exhibit distinct styles, suggesting a distribution gap between the original labels and the target-model outputs. In particular, GPT-style~\citep{bubeck2023gpt4} captions tend to be more literal and position-aware (\eg, ``\textit{in the center of the table}''), whereas the Qwen-series MLLMs~\citep{bai2025qwen3vl} produce captions that are more holistic, semantically enriched, and interpretive (\eg, ``\textit{creatively decorated donuts}''). Such a stylistic mismatch can introduce inconsistent supervision during training, and the resampled distribution better aligns with the pretrained data distribution of models to quantize.

\textbf{Effectiveness of SRL.}
We further evaluate quantized models trained on datasets produced by different labelers in Section~\ref{sec:ablation_study}. The results show that self-generated labels consistently achieve the best performance, outperforming both a stronger closed-source labeler and a homogeneous larger labeler. Appendix~\ref{app:cr_srl_analysis} further compares target-model likelihood, vocabulary coverage, response length, formatting, and semantic similarity on paired cases. These analyses support target-distribution alignment beyond a single stylistic factor and characterize the limits of cross-model label transfer.

\subsection{QAT with Progressive Block Activation}\label{sec:pba}

\begin{figure}[t]
\centering
\begin{minipage}[t]{0.45\linewidth}
    \centering
    \resizebox{\linewidth}{!}{\input{figures/prelim_hellaswag_vs_iter}}
    \subcaption{Language modeling capability.}
    \label{fig:hellaswag_performance}
\end{minipage}
\begin{minipage}[t]{0.05\linewidth}
    \mbox{} %
\end{minipage}
\begin{minipage}[t]{0.45\linewidth}
    \centering
    \resizebox{\linewidth}{!}{\input{figures/prelim_ai2d_vs_iter}}
    \subcaption{MM understanding capability.}
    \label{fig:ai2d_performance}
\end{minipage}
\begin{minipage}[t]{0.05\linewidth}
    \mbox{} %
\end{minipage}
\caption{Accuracy of \texttt{Qwen3-VL-2B-Instruct} W4A8 checkpoints on HellaSwag \citep{zellers2019hellaswag} (left) and AI2D \citep{kembhavi2016ai2d} (right) benchmarks over training iterations.}
\label{fig:prelim_two_benchmarks}
\end{figure}
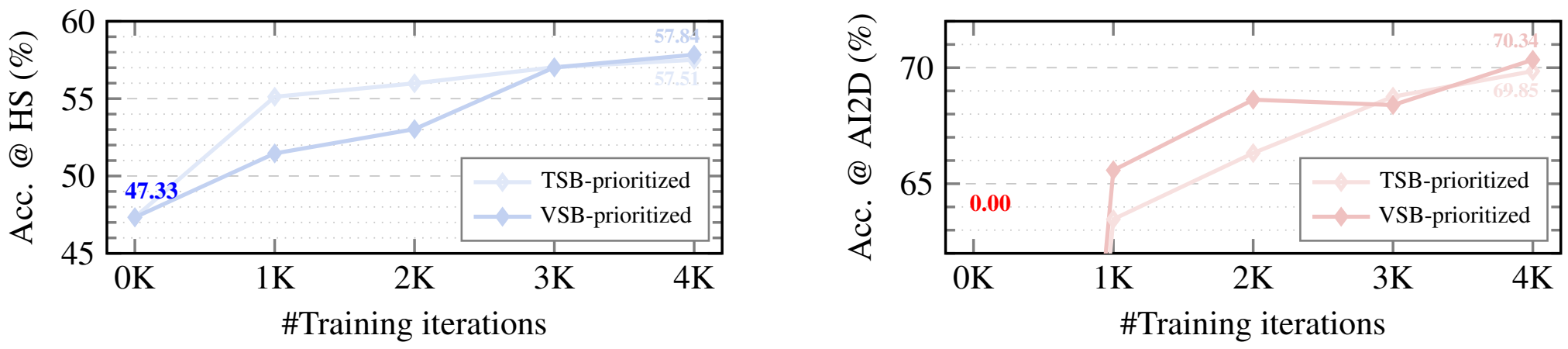

\begin{wrapfigure}{r}{0.4\linewidth}
    \centering
    \vspace{-1.8em}
    \includegraphics[width=\linewidth]{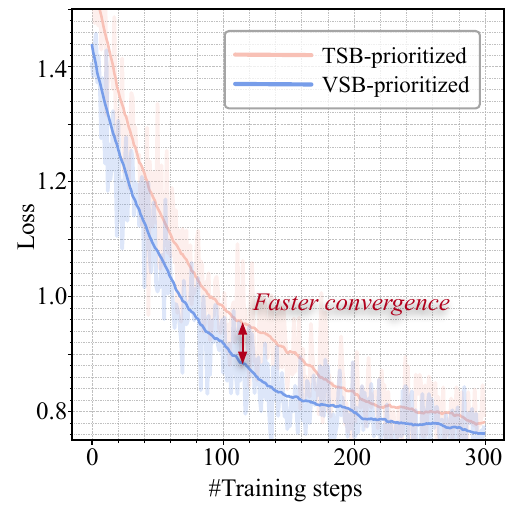}
    \vspace{-1em}
    \caption{Loss trajectories of TSB/VSB-prioritized QAT over the first 300 steps.}
    \vspace{-1.5em}
    \label{fig:preliminary}
\end{wrapfigure}

\textbf{Motivation.} Quantization degrades both MM understanding and language modeling in MLLMs, making QAT substantially more challenging than for text-only LLMs~\citep{liu2023llm,chen2025efficientqat}. Naive QAT attempts to recover these two capabilities jointly, yet yields suboptimal results (see Section~\ref{sec:experiments} for details). 
We therefore decouple MM understanding and language modeling during training via the proposed PBA mechanism. 
As illustrated in Figure~\ref{fig:prelim_two_benchmarks}, initial evaluations of the round-to-nearest baseline (0-th iteration) reveal a stark performance disparity: while the quantized backbone retains a non-trivial linguistic prior (47.33\% accuracy on HellaSwag), its MM understanding collapses entirely (0\% on AI2D).
This phenomenon suggests that cross-modal alignment is significantly more vulnerable to quantization noise than the model's intrinsic generative inertia.
Since MM understanding is the prerequisite anchor for grounded reasoning, we propose to prioritize the restoration of the MM understanding path before progressively refining the more text-sensitive modules.

\begin{figure}[t]
    \centering
    \includegraphics[width=\linewidth]{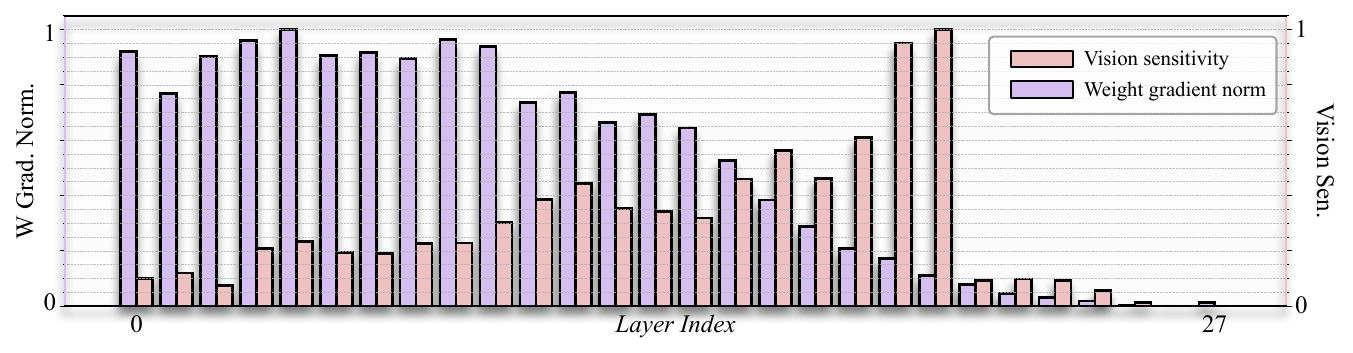}
    \caption{Normalized vision sensitivity $\vec{\tau}$ and weight gradient norm $\lVert \mathbf{G}_w\rVert_{\ell_1}$ across decoding layers.}
    \label{fig:vision_sensitivity}
\end{figure}

\textbf{Vision sensitivity guided PBA mechanism.} 
We propose PBA to prioritize quantization recovery for the vision-sensitive blocks (VSBs) in the LLM backbone.
To identify VSBs, we introduce a vision-sensitivity factor $\vec{\tau}$. 
Specifically, for the $i$-th block, we denote the output gradient as $\mathbf{G}^i=\left[\mathbf{G}^i_v;\ \mathbf{G}^i_t\right]\in\mathbb{R}^{K\times N}$, where $\mathbf{G}^i_v\in\mathbb{R}^{K_v\times N}$ and $\mathbf{G}^i_t\in\mathbb{R}^{K_t\times N}$ correspond to the gradients of vision and text tokens, respectively.
We define the vision-sensitivity factor $\tau^i$ as the ratio between the average token-wise gradient magnitude over vision tokens and that over all tokens, \ie,
\begin{equation}\label{eq:vision_sensitivity}
    \tau^i
    = \frac{K}{K_v} \frac{\sum_{k=1}^{K_v}\left\lVert \mathbf{G}^i_v[k, :]\right\rVert_{\ell_1}}{\sum_{j=1}^{K}\left\lVert \mathbf{G}^i[j, :]\right\rVert_{\ell_1}}.
\end{equation}
Figure~\ref{fig:vision_sensitivity} visualizes $\vec{\tau}=\left[\tau^0,\ \tau^1,\ \ldots,\ \tau^{L-1}\right]$ as \textcolor{myred}{red} bars for \texttt{Qwen3-VL-2B-Instruct} with $L=28$ decoding layers. We observe that VSBs are concentrated in the middle layers, whereas text-sensitive blocks (TSBs) are more prevalent in both shallow and deep layers. Accordingly, we schedule the quantization activation points such that VSBs are quantized first, and TSBs are gradually enabled later in training. 
This process is illustrated in Figure~\ref{fig:overview} (right).

\textbf{Effectiveness of $\tau$.}
We demonstrate the effectiveness of the proposed vision-sensitivity factor $\vec{\tau}$ from two perspectives: convergence behavior and checkpoint performance. We first compare the loss trajectories of the first 300 training steps under TSB-prioritized and VSB-prioritized schedules in Figure~\ref{fig:preliminary}. 
The VSB-prioritized schedule converges faster, indicating that prioritizing multimodal understanding is beneficial for optimization. 
We further report checkpoint performance from 0K to 4K iterations in Figure~\ref{fig:prelim_two_benchmarks}. 
As shown in Figure~\ref{fig:hellaswag_performance}, the VSB-prioritized schedule restores language modeling more slowly at early stages than the TSB-prioritized one, but surpasses it by 4K iterations. 
Meanwhile, in Figure~\ref{fig:ai2d_performance}, the VSB-prioritized schedule recovers MM understanding substantially faster and outperforms the TSB-prioritized schedule at nearly all checkpoints. Overall, $\vec{\tau}$-guided PBA effectively prioritizes the recovery of MM understanding over language modeling, leading to faster early-stage convergence and better final performance.

\begin{figure}[t]
    \centering
    \includegraphics[width=\linewidth]{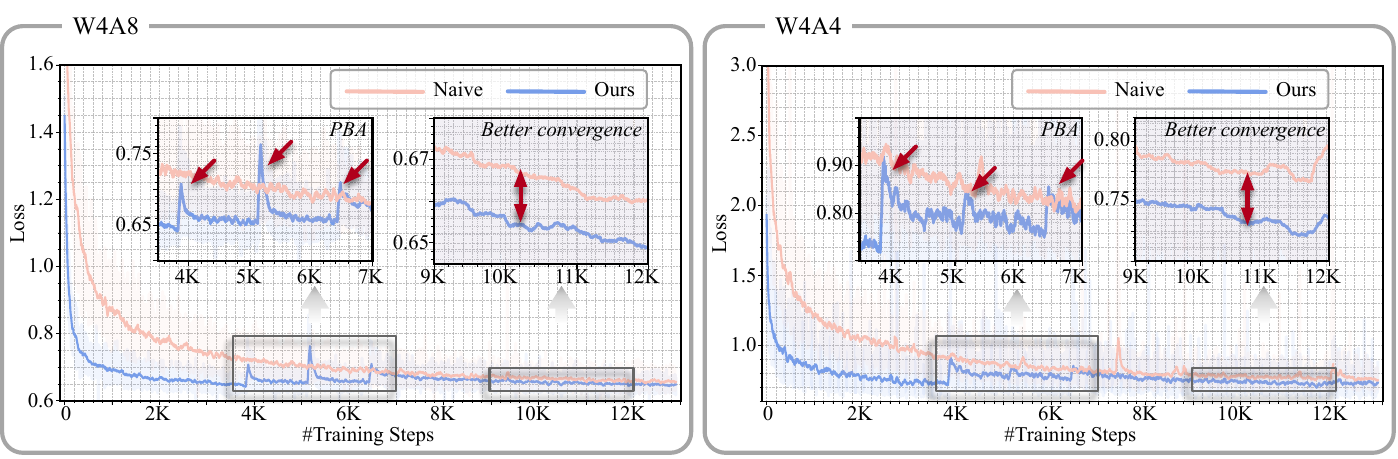}
    \caption{Training loss comparison of \texttt{Qwen3-VL-2B-Instruct} between naive QAT and \method under W4A8 and W4A4. PBA exhibits faster warm-up convergence toward the pre-trained level, rapid recovery during PBA, and better final convergence.}
    \label{fig:loss}
\end{figure}

\textbf{Effectiveness of PBA.}
We further validate the effectiveness of PBA by comparing the full training trajectories in Figure~\ref{fig:loss} against a naive QAT baseline. 
Under both W4A8 and W4A4 settings, PBA approaches the pre-trained loss level more rapidly during the warm-up stage. Once new block quantization is activated, the loss quickly descends below the naive baseline, despite tolerable spikes naturally introduced by the staged quantization. In the subsequent post-hoc recovery stage, PBA attains a better optimum as well. This advantage is further corroborated by the performance gains reported in Section~\ref{sec:experiments}.

%% file: figures/prelim_hellaswag_vs_iter.tex
\begin{tikzpicture} 
\begin{axis}[ 
    width=8cm, %
    height=4cm, %
    xmin=0.0, xmax=4.0, %
    enlarge x limits=0.05,
    ymin=45.00, ymax=60.00, %
    xtick={0,1,2,3,4}, %
    minor x tick num=0,
    xticklabels={0K,1K,2K,3K,4K}, %
    ytick={45,50,55,60}, %
    minor y tick num=4,
    yticklabels={45,50,55,60}, %
    xlabel={\#Training iterations}, %
    ylabel={Acc. @ HS (\%)}, %
    grid=both, %
    ymajorgrids=true, %
    yminorgrids=true, %
    xmajorgrids=false, %
    xminorgrids=false, %
    major grid style={gray!45,dashed}, %
    minor grid style={gray!45,dotted}, %
    axis line style={thick}, %
    tick style={thick}, %
    label style={font=\normalsize}, %
    ticklabel style={font=\normalsize}, %
    legend style={ 
        at={(0.98,0.05)}, %
        anchor=south east, %
        draw=gray, 
        line width=0.6pt,
        fill=white, 
        font=\scriptsize %
    } 
] 

\addplot[ 
    myblue!50, 
    solid,
    line width=1.2pt, 
    mark=diamond, 
    mark size=2.0pt
] coordinates { 
    (0,47.33) 
    (1,55.12) 
    (2,55.99) 
    (3,57.03) 
    (4,57.51) 
}; 
\node[below, xshift=-5pt, yshift=0pt, font=\tiny, text=myblue!50] at (axis cs:4,57.51) {\textbf{57.51}};
\addlegendentry{TSB-prioritized} 

\addplot[ 
    myblue,
    solid,
    line width=1.2pt, 
    mark=diamond*, 
    mark size=2.0pt
] coordinates { 
    (0,47.33) 
    (1,51.46) 
    (2,53.03) 
    (3,57.02) 
    (4,57.84)
};
\node[above, xshift=-5pt, yshift=0pt, font=\tiny, text=myblue] at (axis cs:4,57.84) {\textbf{57.84}};
\node[above, xshift=5pt, yshift=2pt, font=\scriptsize, text=blue] at (axis cs:0,47.33) {\textbf{47.33}};
\addlegendentry{VSB-prioritized} 

\end{axis} 
\end{tikzpicture} 

%% file: figures/prelim_ai2d_vs_iter.tex
\begin{tikzpicture} 
\begin{axis}[ 
    width=8cm, %
    height=4cm, %
    xmin=0.0, xmax=4.0, %
    enlarge x limits=0.05,
    ymin=62, ymax=72, %
    xtick={0,1,2,3,4}, %
    minor x tick num=0,
    xticklabels={0K,1K,2K,3K,4K}, %
    ytick={60,65,70,75}, %
    minor y tick num=4,
    yticklabels={60,65,70,75}, %
    xlabel={\#Training iterations}, %
    ylabel={Acc. @ AI2D (\%)}, %
    grid=both, %
    ymajorgrids=true, %
    yminorgrids=true, %
    xmajorgrids=false, %
    xminorgrids=false, %
    major grid style={gray!45,dashed}, %
    minor grid style={gray!45,dotted}, %
    axis line style={thick}, %
    tick style={thick}, %
    label style={font=\normalsize}, %
    ticklabel style={font=\normalsize}, %
    legend style={ 
        at={(0.98,0.05)}, %
        anchor=south east, %
        draw=gray, 
        line width=0.6pt,
        fill=white, 
        font=\scriptsize %
    } 
] 

\addplot[ 
    myred!50, 
    solid,
    line width=1.2pt, 
    mark=diamond, 
    mark size=2.0pt
] coordinates { 
    (0, 0.00)
    (1,63.47) 
    (2,66.32) 
    (3,68.75) 
    (4,69.85)
};
\node[below, xshift=-5pt, yshift=0pt, font=\tiny, text=myred!50] at (axis cs:4,69.85) {\textbf{69.85}};
\addlegendentry{TSB-prioritized} 

\addplot[ 
    myred,
    solid,
    line width=1.2pt, 
    mark=diamond*, 
    mark size=2.0pt
] coordinates { 
    (0, 0.00)
    (1,65.58) 
    (2,68.62) 
    (3,68.39) 
    (4,70.34)
};
\node[above, xshift=-5pt, yshift=0pt, font=\tiny, text=myred] at (axis cs:4,70.34) {\textbf{70.34}};
\node[above, xshift=5pt, yshift=2pt, font=\scriptsize, text=red] at (axis cs:0,63) {\textbf{0.00}};
\addlegendentry{VSB-prioritized} 

\end{axis} 
\end{tikzpicture} 

%% file: docs/experiments.tex
\section{Experiments}\label{sec:experiments}

\subsection{Implementation Details}\label{sec:implementation_details}

\textbf{Models and datasets.} We conduct experiments on small-scale models, including \texttt{Qwen3-VL-2B-Instruct}~\citep{bai2025qwen3vl}, \texttt{Qwen2.5-VL-3B-Instruct}~\citep{bai2025qwen25vl}, and \texttt{Qwen2-VL-2B-Instruct}~\citep{wang2024qwen2vl}. To demonstrate the generalization of \method, we further extend our evaluation to a large model, \texttt{Qwen3-VL-8B-Instruct}. The training data are adapted from \texttt{LLaVA-680K}~\citep{liu2024llava1.5}.

\textbf{Baselines.} We compare the proposed \method with strong PTQ and QAT baselines. Specifically, we consider outlier-smoothing methods, including MBQ~\citep{li2025mbq} and SmoothQuant~\citep{xiao2023smoothquant}, the rotation-assisted approach MQuant~\citep{yu2025mquant}, as well as the QAT solution LLM-QAT~\citep{liu2023llm}. PTQ baselines use their official implementations and paper configurations. We adapt LLM-QAT to multimodal inputs using the same training data, budget, and optimization settings as \method; the marked W4A4 cases retain 8-bit \texttt{down\_proj} activations. Further implementation details are provided in
Appendix~\ref{app:more_impl}.

\textbf{Evaluation.} To evaluate quantized models, we report results on a diverse suite of benchmarks built on LMMs-Eval~\citep{zhang2024lmms-eval}: MME~\citep{fu2023mme} for comprehensive multi-modal evaluation; AI2D~\citep{kembhavi2016ai2d} and TextVQA for text recognition; ChartQA~\citep{masry2022chartqa} for chart parsing; DocVQA~\citep{mathew2021docvqa} for document understanding; MMMU~\citep{yue2023mmmu} and ScienceQA~\citep{lu2022scienceqa} for visual reasoning; and SeedBench~\citep{li2023seedbench} for spatial–temporal image/video understanding.

\begin{table}[t]
    \centering
    \caption{Weight-activation quantization results for Qwen VL series. The headers represent model architectures, bitwidth, quantization methods, AI2D, ChartQA, DocVQA, MMMU, ScienceQA, SeedBench, TextVQA, average accuracy, and MME scores, respectively. The superscript $\dag$ indicates \texttt{down\_proj} activations are quantized to 8-bit under W4A4 settings.}
    \resizebox{\linewidth}{!}{\input{tables/exp_main}}
    \label{tab:exp_main}
\end{table}

\subsection{Performance Analysis}\label{sec:performance_analysis}

\subsubsection{Performance under W4A8 \& W4A4 quantization.}
We report the performance of quantized models on a diverse set of MM benchmarks in Table~\ref{tab:exp_main}. Overall, \method consistently outperforms both PTQ and QAT baselines across all model backbones and quantization settings, with the advantage becoming especially significant in the more aggressive W4A4 regime.

Under W4A8 quantization, \method preserves the performance of the full-precision models remarkably well. Across all three backbones, the average accuracy of \method remains within approximately one point of the BF16 counterparts (i.e., around $\sim 1\%$ absolute degradation), indicating near-lossless quantization in this setting. In contrast, existing PTQ and QAT baselines exhibit noticeably larger drops on multiple benchmarks, even when their performance is competitive on a subset of metrics.

The performance gap becomes much larger under W4A4 quantization. In this regime, PTQ baselines (e.g., MBQ and MQuant) largely collapse, with near-zero or extremely low scores on multiple benchmarks, indicating that they fail to maintain usable multi-modal reasoning and perception abilities at 4-bit activation quantization. The LLM-QAT baseline is substantially more robust than PTQ, but still suffers severe degradation; in particular, it can only maintain reasonable performance when \texttt{down\_proj} activations are kept at 8-bit. Even with this relaxation, its performance remains far below \method in most cases.

\subsubsection{Performance under W8A8 quantization.}

\begin{table}[t]
    \centering
    \caption{W8A8 quantization results on three model backbones. Values in \textcolor{myred}{red} denote the corresponding full-precision (BF16) performance.}
    \resizebox{0.8\linewidth}{!}{\input{tables/exp_w8a8}}
    \label{tab:exp_w8a8}
\end{table}

Under the relatively mild W8A8 setting, both PTQ approaches and \method achieve performance comparable to BF16 across all three backbones, indicating that strong PTQ methods are already sufficient in this regime and that the extra training cost of QAT offers limited cost-effectiveness. In contrast, the advantage of \method becomes much more pronounced in lower-bit settings (\eg, W4A8/W4A4) as demonstrated in Table~\ref{tab:exp_main}, making it more suitable for aggressive low-bit quantization where performance preservation is substantially more challenging.

\subsubsection{Compatibility on weight-only quantization.}
Results in Table~\ref{tab:exp_wonly} show that \method generalizes well beyond W/A quantization and remains effective in weight-only (group-wise) quantization settings. Specifically, under both W4g128 and the more aggressive W3g128 configurations, \method consistently outperforms the PTQ baseline on all benchmarks and yields clear gains in both average accuracy and MME. The advantage is particularly significant at W3g128, where MBQ suffers severe degradation while \method still preserves strong MM performance, demonstrating the robustness and generalizability of our method to low-bit weight-only quantization.

\begin{table}[t]
    \centering
    \caption{Weight-only quantization results of MBQ and \method under W4g128 and W3g128 settings.}
    \resizebox{0.9\linewidth}{!}{\input{tables/exp_wonly}}
    \label{tab:exp_wonly}
\end{table}

\subsubsection{Scale to large models.}

\begin{table}[t]
    \centering
    \caption{W/A quantization results on \texttt{Qwen3-VL-\textcolor{myred}{8B}-Instruct}.}
    \resizebox{0.9\linewidth}{!}{\input{tables/exp_8b}}
    \label{tab:exp_8b}
\end{table}

Results on \texttt{Qwen3-VL-8B-Instruct} in Table~\ref{tab:exp_8b} show that \method generalizes well beyond the small-sized models targeted in our main experiments. It remains highly effective on this large backbone, consistently outperforming the PTQ baseline under both W4A8 and W4A4 quantization. In particular, \method preserves near-full-precision performance at W4A8 and maintains strong MM capability even under the aggressive W4A4 setting, while MBQ collapses. These results demonstrate the strong scalability and generalization ability of \method across model sizes.

\subsubsection{Mixed-precision quantization.}

As pointed out in QAT scaling law~\citep{chen2025qat-scaling-law}, persistent outliers in \texttt{down\_proj} activations are a major bottleneck for low-bit QAT. Consistent with this observation, our mixed-precision results in Table~\ref{tab:exp_mix_precision} show that keeping \texttt{down\_proj} activations at 8-bit (W4A4$^\dag$) consistently improves performance over full W4A4 quantization.

\begin{wraptable}{r}{0.55\linewidth}
    \centering
    \vspace{-2.5em}
    \caption{Mixed-precision quantization results on three backbones. The superscript $\dag$ denotes the W4A4$^\dag$ setting, where \texttt{down\_proj} activations are kept at 8-bit.}
    \resizebox{\linewidth}{!}{\input{tables/exp_mix_precision}}
    \vspace{-1em}
    \label{tab:exp_mix_precision}
\end{wraptable}
However, the gain varies substantially across backbones (\textit{e.g.}, +1.44, +9.39, and +3.21 Avg. points), suggesting that the benefit of mixed-precision design is relatively model-dependent. Therefore, mixed-precision quantization likely requires a more fine-grained, backbone-aware treatment. Here we provide only preliminary evidence and leave a systematic analysis to future work.

\subsection{Ablation Study}\label{sec:ablation_study}
\subsubsection{Step-by-step ablation analysis.}
\begin{figure}[ht]
    \centering
    \begin{minipage}[t]{0.48\linewidth}
        \centering
        \resizebox{\linewidth}{!}{\input{figures/abl_step_by_step_w4a8}}
        \subcaption{W4A8.}
    \end{minipage}
    \begin{minipage}[t]{0.02\linewidth}
        \mbox{} %
    \end{minipage}
    \begin{minipage}[t]{0.48\linewidth}
        \centering
        \resizebox{\linewidth}{!}{\input{figures/abl_step_by_step_w4a4}}
        \subcaption{W4A4.}
    \end{minipage}
    \begin{minipage}[t]{0.02\linewidth}
        \mbox{} %
    \end{minipage}
    \caption{The step-by-step ablation studies on SRL and PBA.}
    \label{fig:abl_step_by_step}
\end{figure}
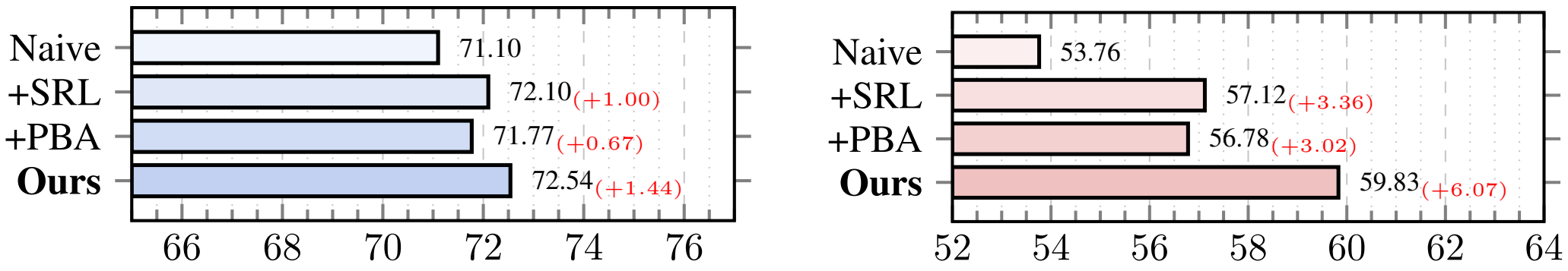
We conduct step-by-step ablation studies on \texttt{Qwen3-VL-2B-Instruct} under W4A8 and W4A4 settings in Figure~\ref{fig:abl_step_by_step}. Starting from the naive QAT baseline, the proposed SRL scheme improves the average accuracy by +1.00 and +3.36 under W4A8 and W4A4, respectively, indicating that refined supervision alone already provides meaningful gains. Applying PBA to the same baseline yields gains of +0.67 and +3.02, confirming that the staged quantization strategy also contributes independently by better preserving visual-sensitive layers. Combining both components further increases the average accuracy by +1.44 and +6.07, respectively, achieving the best performance in both settings and demonstrating that SRL and PBA are complementary.

\subsubsection{Ablation on block activation metric.}
\begin{wraptable}{r}{0.3\linewidth}
    \centering
    \vspace{-1em}
    \caption{Ablation studies on block activation metric}
    \resizebox{\linewidth}{!}{\input{tables/abl_metric}}
    \vspace{-1em}
    \label{tab:abl_metric}
\end{wraptable}

Table~\ref{tab:abl_metric} compares different metrics and priors for block quantization activation priority. A random selection yields an average accuracy of 71.88, serving as a baseline. We evaluate the performance by taking the weight gradients norm $\lVert\mathbf{G}_w\rVert_{\ell_1}$ as an activation metric. As plotted in Figure~\ref{fig:vision_sensitivity}, early layers have larger $\lVert\mathbf{G}_w\rVert_{\ell_1}$ than that of deep layers. Therefore, $\lVert\mathbf{G}_w\rVert_{\ell_1}$ (Higher) ($\lVert\mathbf{G}_w\rVert_{\ell_1}$ (Lower)) in Table~\ref{tab:abl_metric} activates layers from early to deep layers (from deep to shallow layers) approximately. With our vision-sensitivity factor $\tau$, the lower-prior setting (TSB-prioritized) is slightly weaker, whereas prioritizing higher-$\tau$ blocks (VSB-prioritized) achieves the best result. Overall, these results validate that $\tau$ is a more effective criterion for PBA, and that allocating higher priority to the most VSBs is crucial for maximizing performance.

\subsubsection{Ablation on PBA scheduler.}

As shown in Table~\ref{tab:abl_scheduler}, we test the PBA scheduler across 3 different annealing functions, which are \textit{Cosine}, \textit{Linear}, and \textit{Logarithm}, respectively. Taking \texttt{Qwen3-VL-2B-Instruct} with 28 decoding layers as an example, we employ the \textit{Logarithm} functionalized scheduler, which activates 50\%, 75\%, 88\%, and 100\% layers as training progresses. Training iterations within each stage are empirically set to $\sim$10\% of total iterations, with the remaining budgets assigned to warmup and post-hoc recovery. Our ablation study demonstrates that \textit{Logarithm} scheduler, which activates more blocks at early stages, yields better performance than \textit{Cosine} and \textit{Linear} functions.

\subsubsection{Ablation on block granularity.}

\begin{table}[t]
    \centering
    \caption{Ablation studies on SRL labeler, PBA scheduler, and block granularity.}
    \begin{minipage}{0.30\linewidth}
        \centering
        \subcaption{PBA scheduler.}
        \resizebox{0.92\linewidth}{!}{\input{tables/abl_scheduler}}
        \label{tab:abl_scheduler}
    \end{minipage}
    \hfill
    \begin{minipage}{0.32\linewidth}
        \centering
        \subcaption{Block granularity}
        \resizebox{0.87\linewidth}{!}{\input{tables/abl_block_granularity}}
        \label{tab:abl_block_granularity}
    \end{minipage}
    \hfill
    \begin{minipage}{0.35\linewidth}
        \centering
        \subcaption{SRL labeler.}
        \resizebox{0.89\linewidth}{!}{\input{tables/abl_labeler}}
        \label{tab:abl_labeler}
    \end{minipage}
\end{table}

We explore how the block splitting strategy affects performance. Table~\ref{tab:abl_block_granularity} shows that treating each decoding layer as a single activation unit ($1\times\theta_\text{DecLayer}$) yields the best accuracy. 
$0.5\times\theta_\text{DecLayer}$ treats the \texttt{attn} and \texttt{ffn} modules as separate units, which can trigger asynchronous quantization activation within the same layer and thus hurts performance. 
$2\times\theta_\text{DecLayer}$ leads to severe degradation, suggesting that overly coarse granularity has detrimental side effects. 
One reason is that coarse units may mix blocks with substantially different vision sensitivity, forcing them to share the same activation schedule. 
As a result, quantization may be enabled too early for less robust components or too late for truly vision-critical ones, weakening the intended MM-understanding-first recovery. 
This is consistent with Figure~\ref{fig:vision_sensitivity}, where consecutive layers can exhibit large variations in visual sensitivity.

\subsubsection{Ablation on SRL labeler.}
We conduct experiments on \texttt{Qwen3-VL-2B-Instruct} with dataset labeled by a closed-source stronger model \texttt{GPT-4}~\citep{liu2024llava1.5}, self (\texttt{Qwen3-VL-2B-Instruct}) and homogeneous larger model (Homo. 8B$^\ddag$, \texttt{Qwen3-VL-8B-Instruct}). Results in Table~\ref{tab:abl_labeler} show that both self and homogeneous larger models outperform the distribution-mismatched \texttt{GPT-4} labeler by a large margin. Furthermore, self-annotated datasets are better aligned with the target distribution and yield the best results. The extended labeler comparison and distribution diagnostics in Appendix~\ref{app:cr_srl_analysis} show that self-labels outperform same-family and cross-family alternatives. Per-model relabeling remains preferable when its generation cost is acceptable.

\subsection{Hardware Efficiency on Edge Accelerator}\label{sec:accelerator}
To assess edge deployment, we evaluate a silicon-proven TSMC 22\,nm ASIC using measured on-chip latency and average power, off-chip DRAM simulation, and an end-to-end cycle-accurate accelerator model. The boundary covers the accelerator and DRAM accesses. Across three backbones, \method achieves up to $3.64\times$ prefill speedup and $5.64\times$ energy reduction. Appendix~\ref{app:cr_measurement} documents the setup and measurement methodology; Appendices~\ref{app:more_hardware_results} and~\ref{app:cr_w8a8} report results across precisions and token configurations.

%% file: tables/exp_main.tex
\begin{tabular}{
    >{\centering\arraybackslash}m{0.6cm}
    >{\centering\arraybackslash}m{1.3cm}
    >{\centering\arraybackslash}m{1.8cm}
    *{7}{>{\centering\arraybackslash}m{1.0cm}}
    >{\centering\arraybackslash}m{1.4cm}
    >{\centering\arraybackslash}m{1.4cm}
}
\toprule
\textbf{Arch} & \textbf{\#BW} & \textbf{Method} & \textbf{AI2D} & \textbf{ChQ} & \textbf{DocV} & \textbf{M3U} & \textbf{SciQ} & \textbf{Seed} & \textbf{TxtV} & \textbf{Avg.} ($\uparrow$) & \textbf{MME} \\
\midrule
\multirow[c]{10}{*}{\rotatebox[origin=c]{90}{Qwen2-VL-2B}} & BF16 & - & 70.04 & 72.16 & 89.40 & 39.89 & 74.23 & 69.98 & 79.38 & 70.73 & 1870.19 \\
\cmidrule(l){2-12}
& \multirow[c]{5}{*}{W4A8} & SmoothQ & 62.66 & 53.68 & 76.59 & 34.00 & 66.47 & 63.20 & 69.17 & 60.82 & 1736.61 \\
& & MBQ & 65.12 & 62.08 & 79.76 & 35.22 & 68.99 & 64.64 & 71.91 & 63.96 & 1810.68 \\
& & MQuant & 67.36 & \textbf{70.44} & 87.39 & 38.33 & -- & -- & 77.72 & -- & -- \\
& & LLM-QAT & 66.94 & 68.20 & 87.14 & 35.56 & 69.37 & 68.27 & 77.94 & 67.63 & \textbf{1931.70} \\
& & \bg \textbf{\method} & \bg \textbf{69.30} & \bg 70.32 & \bg \textbf{88.10} & \bg \textbf{39.00} & \bg \textbf{72.79} & \bg \textbf{69.19} & \bg \textbf{78.60} & \bg \textbf{69.61} & \bg 1830.28 \\
\cmidrule(l){2-12}
& \multirow[c]{4}{*}{W4A4} & MBQ & 2.17 & 0.00 & 0.00 & 23.67 & 1.53 & 8.38 & 0.00 & 5.11 & 55.41 \\
& & MQuant & 11.30 & 0.36 & 0.16 & 12.56 & -- & -- & 0.18 & -- & -- \\
& & LLM-QAT$^\dag$ & 29.95 & 39.68 & 66.74 & 27.56 & 43.10 & 40.68 & 69.94 & 45.38 & 1532.13 \\
& & \bg \textbf{\method} & \bg \textbf{40.97} & \bg \textbf{47.32} & \bg \textbf{72.27} & \bg \textbf{30.11} & \bg \textbf{56.99} & \bg \textbf{61.07} & \bg \textbf{72.40} & \bg \textbf{54.45} & \bg \textbf{1611.94} \\
\midrule
\multirow[c]{8}{*}{\rotatebox[origin=c]{90}{Qwen2.5-VL-3B}} & BF16 & - & 78.63 & 83.32 & 92.50 & 47.00 & 83.33 & 70.94 & 78.61 & 76.33 & 2149.00 \\
\cmidrule(l){2-12}
& \multirow[c]{4}{*}{W4A8} & SmoothQ & 71.83 & 74.36 & 79.65 & 40.44 & 73.92 & 66.90 & 63.91 & 67.29 & 2063.23 \\
& & MBQ   & 73.87 & 79.24 & 88.44 & 40.44 & 77.18 & 69.27 & 74.00 & 71.78 & 2049.63 \\
& & LLM-QAT   & 54.86 & 72.76 & 62.58 & 33.00 & 53.24 & 49.07 & 71.35 & 56.69 & 1939.38 \\
& & \bg \textbf{\method} & \bg \textbf{77.27} & \bg \textbf{82.68} & \bg \textbf{91.26} & \bg \textbf{44.11} & \bg \textbf{81.61} & \bg \textbf{70.80} & \bg \textbf{78.21} & \bg \textbf{75.13} & \bg \textbf{2083.30} \\
\cmidrule(l){2-12}
& \multirow[c]{3}{*}{W4A4} & MBQ   & 0.94 & 0.00 & 0.51 & 24.33 & 0.87 & 5.35 & 0.02 & 4.57 & 11.24 \\
& & LLM-QAT$^\dag$   & 31.22 & 44.96 & \textbf{66.40} & 27.56 & 45.84 & 24.99 & 64.75 & 43.67 & 1090.88 \\
& & \bg \textbf{\method} & \bg \textbf{48.48} & \bg \textbf{51.00} & \bg 64.53 & \bg \textbf{30.67} & \bg \textbf{61.64} & \bg \textbf{61.56} & \bg \textbf{67.65} & \bg \textbf{55.08} & \bg \textbf{1520.12} \\
\midrule
\multirow[c]{8}{*}{\rotatebox[origin=c]{90}{Qwen3-VL-2B}} & BF16  & - & 73.25 & 79.56 & 92.70 & 38.78 & 79.58 & 72.26 & 79.26 & 73.63 & 2029.18 \\
\cmidrule(l){2-12}
& \multirow[c]{4}{*}{W4A8} & SmoothQ & 48.64 & 61.08 & 80.84 & 33.00 & 40.18 & 64.91 & 64.82 & 56.21 & 1834.91 \\
& & MBQ   & 50.62 & 67.44 & 84.20 & 32.78 & 50.51 & 65.75 & 69.40 & 60.10 & 1758.88 \\
& & LLM-QAT   & 60.52 & 65.88 & 86.11 & 38.89 & 69.02 & 67.45 & 73.71 & 65.94 & \textbf{1962.77} \\
& & \bg \textbf{\method} & \bg \textbf{71.34} & \bg \textbf{75.44} & \bg \textbf{91.10} & \bg \textbf{41.22} & \bg \textbf{78.73} & \bg \textbf{71.90} & \bg \textbf{78.08} & \bg \textbf{72.54} & \bg 1962.04 \\
\cmidrule(l){2-12}
& \multirow[c]{3}{*}{W4A4} & MBQ   & 5.18 & 0.00 & 0.10 & 25.56 & 6.65 & 7.03 & 0.08 & 6.37 & 0.00 \\
& & LLM-QAT$^\dag$   & 19.95 & 28.68 & 25.39 & 26.33 & 30.91 & 15.94 & 43.82 & 27.29 & 764.33 \\
& & \bg \textbf{\method} & \bg \textbf{54.86} & \bg \textbf{54.12} & \bg \textbf{73.23} & \bg \textbf{33.11} & \bg \textbf{64.84} & \bg \textbf{67.47} & \bg \textbf{71.18} & \bg \textbf{59.83} & \bg \textbf{1652.63} \\
\bottomrule
\end{tabular}%

%% file: tables/exp_w8a8.tex
\begin{tabular}{
    >{\centering\arraybackslash}m{1.8cm}
    *{7}{>{\centering\arraybackslash}m{1.0cm}}
    >{\centering\arraybackslash}m{1.4cm}
    >{\centering\arraybackslash}m{1.4cm}
}
\toprule
\textbf{Method} & \textbf{AI2D} & \textbf{ChQ} & \textbf{DocV} & \textbf{M3U} & \textbf{SciQ} & \textbf{Seed} & \textbf{TxtV} & \textbf{Avg.} ($\uparrow$) & \textbf{MME} \\

\midrule
\multicolumn{8}{l}{\texttt{\ Qwen2-VL-2B-Instruct}} & \textcolor{myred}{70.73} & \textcolor{myred}{1870.19} \\
\midrule
MBQ & \textbf{69.88} & \textbf{72.28} & 86.89 & \textbf{39.33} & \textbf{74.09} & 68.82 & 77.13 & 69.77 & \textbf{1884.59} \\
\bg \textbf{\method} & \bg 69.66 & \bg 72.08 & \bg \textbf{88.91} & \bg 38.89 & \bg 73.21 & \bg \textbf{69.88} & \bg \textbf{79.34} & \bg \textbf{70.28} & \bg 1864.33 \\

\midrule
\multicolumn{8}{l}{\texttt{\ Qwen2.5-VL-3B-Instruct}} & \textcolor{myred}{76.33} & \textcolor{myred}{2149.00} \\
\midrule
MBQ & 78.27 & 83.00 & \textbf{92.69} & 46.78 & \textbf{82.55} & \textbf{71.12} & \textbf{78.60} & 76.14 & \textbf{2156.88} \\
\bg \textbf{\method} & \bg \textbf{78.30} & \bg \textbf{83.52} & \bg 92.25 & \bg \textbf{48.11} & \bg 82.34 & \bg 71.08 & \bg 78.42 & \bg \textbf{76.29} & \bg 2102.85 \\

\midrule
\multicolumn{8}{l}{\texttt{\ Qwen3-VL-2B-Instruct}} & \textcolor{myred}{73.63} & \textcolor{myred}{2029.18} \\
\midrule
MBQ & \textbf{75.00} & \textbf{79.96} & \textbf{92.80} & 38.99 & \textbf{79.56} & 71.98 & \textbf{79.35} & \textbf{73.95} & \textbf{2009.60} \\
\bg \textbf{\method} & \bg 72.25 & \bg 78.64 & \bg 92.35 & \bg \textbf{39.22} & \bg 76.75 & \bg \textbf{72.28} & \bg 78.65 & \bg 72.88 & \bg 1996.02 \\

\bottomrule
\end{tabular}%

%% file: tables/exp_wonly.tex
\begin{tabular}{
    >{\centering\arraybackslash}m{1.3cm}
    >{\centering\arraybackslash}m{1.8cm}
    *{7}{>{\centering\arraybackslash}m{1.0cm}}
    >{\centering\arraybackslash}m{1.4cm}
    >{\centering\arraybackslash}m{1.4cm}
}
\toprule
\textbf{\#BW} & \textbf{Method} & \textbf{AI2D} & \textbf{ChQ} & \textbf{DocV} & \textbf{M3U} & \textbf{SciQ} & \textbf{Seed} & \textbf{TxtV} & \textbf{Avg.} ($\uparrow$) & \textbf{MME} \\
\midrule
\multirow[c]{2}{*}{W4g128} & MBQ & 64.35 & 74.88 & 91.23 & 36.67 & 68.57 & 70.22 & 76.94 & 68.98 & 1955.42 \\
& \bg \textbf{\method} & \bg \textbf{71.76} & \bg \textbf{76.44} & \bg \textbf{91.26} & \bg \textbf{38.67} & \bg \textbf{77.08} & \bg \textbf{72.11} & \bg \textbf{77.97} & \bg \textbf{72.18} & \bg \textbf{1958.53} \\
\midrule
\multirow[c]{2}{*}{W3g128} & MBQ & 32.25 & 61.00 & 76.21 & 28.00 & 46.59 & 62.91 & 65.48 & 53.21 & 1350.49 \\
& \bg \textbf{\method} & \bg \textbf{65.90} & \bg \textbf{71.92} & \bg \textbf{88.29} & \bg \textbf{35.67} & \bg \textbf{68.45} & \bg \textbf{70.23} & \bg \textbf{75.60} & \bg \textbf{68.01} & \bg \textbf{1922.82} \\
\bottomrule
\end{tabular}%

%% file: tables/exp_8b.tex
\begin{tabular}{
    >{\centering\arraybackslash}m{1.3cm}
    >{\centering\arraybackslash}m{1.8cm}
    *{7}{>{\centering\arraybackslash}m{1.0cm}}
    >{\centering\arraybackslash}m{1.4cm}
    >{\centering\arraybackslash}m{1.4cm}
}
\toprule
\textbf{\#BW} & \textbf{Method} & \textbf{AI2D} & \textbf{ChQ} & \textbf{DocV} & \textbf{M3U} & \textbf{SciQ} & \textbf{Seed} & \textbf{TxtV} & \textbf{Avg.} ($\uparrow$) & \textbf{MME} \\
\midrule
FP & - & 83.71 & 84.80 & 95.53 & 50.22 & 92.88 & 75.31 & 82.48 & 80.70 & 2381.20 \\
\midrule
\multirow[c]{2}{*}{W4A8} 
& MBQ & 77.62 & 75.40 & 92.71 & 45.78 & 87.31 & 72.55 & 78.00 & 75.62 & 2162.47 \\
& \bg \textbf{\method} & \bg \textbf{82.19} & \bg \textbf{84.12} & \bg \textbf{95.20} & \bg \textbf{49.33} & \bg \textbf{91.44} & \bg \textbf{75.16} & \bg \textbf{81.18} & \bg \textbf{79.80} & \bg \textbf{2325.49} \\
\midrule
\multirow[c]{2}{*}{W4A4} 
& MBQ & 2.20 & 0.00 & 0.03 & 26.22 & 3.56 & 8.30 & 0.00 & 5.76 & 0.00 \\
& \bg \textbf{\method} & \bg \textbf{62.44} & \bg \textbf{57.36} & \bg \textbf{88.22} & \bg \textbf{40.89} & \bg \textbf{76.77} & \bg \textbf{67.53} & \bg \textbf{75.32} & \bg \textbf{66.93} & \bg \textbf{1877.45} \\
\bottomrule
\end{tabular}%

%% file: tables/exp_mix_precision.tex
\begin{tabular}{
    >{\centering\arraybackslash}m{1.2cm}
    >{\centering\arraybackslash}m{1.1cm}
    >{\centering\arraybackslash}m{1.1cm}
    >{\centering\arraybackslash}m{1.15cm}
    >{\centering\arraybackslash}m{1.15cm}
    >{\centering\arraybackslash}m{1.1cm}
    >{\centering\arraybackslash}m{1.1cm}
}
\toprule
\multirow{2}{*}{\textbf{\#BW}}  & \multicolumn{2}{c}{\texttt{Qwen2-VL-2B}} & \multicolumn{2}{c}{\texttt{Qwen2.5-VL-3B}} & \multicolumn{2}{c}{\texttt{Qwen3-VL-2B}} \\
\cmidrule(lr){2-3}\cmidrule(lr){4-5}\cmidrule(lr){6-7}
 & \textbf{Avg.} & $\mathbf{\Delta}$ & \textbf{Avg.} & $\mathbf{\Delta}$ & \textbf{Avg.} & $\mathbf{\Delta}$ \\
\midrule
W4A4$^{\phantom{\dag}}$           & 54.45 & -- & 55.08 & -- & 59.83 & -- \\
\rowbg W4A4$^\dag$  & 55.89 & \textcolor{red}{+1.44} & 64.47 & \textcolor{red}{+9.39} & 63.04 & \textcolor{red}{+3.21} \\
\bottomrule
\end{tabular}

%% file: figures/abl_step_by_step_w4a8.tex
\begin{tikzpicture}
\begin{axis}[
    xbar,                                  %
    width=7cm,
    height=3.5cm,
    xmin=65.0, xmax=77.0,
    xtick={64,66,68,70,72,74,76,78}, %
    minor x tick num=3,
    y dir=reverse,                         %
    ytick={1,2,3,4},
    yticklabels={Naive,+SRL,+PBA,\textbf{Ours}},
    bar width=8pt,
    enlarge y limits=0.3,
    axis line style={thick}, %
    tick style={thick}, %
    ymajorgrids=false, %
    yminorgrids=false, %
    xmajorgrids=true, %
    xminorgrids=true, %
    major grid style={gray!45,dashed}, %
    minor grid style={gray!45,dotted}, %
]

\addplot[
    draw=black,                           
    fill=myblue!25,                        
    line width=1.0pt,
    bar shift=0pt
] coordinates {
    (71.10,1)   
};
\node[anchor=west, xshift=2pt, yshift=0pt, font=\scriptsize] at (axis cs:71.10,1) {71.10};

\addplot[
    draw=black,                           
    fill=myblue!50,                        
    line width=1.0pt,
    bar shift=0pt
] coordinates {
    (72.10,2)
};
\node[anchor=west, xshift=2pt, yshift=-1pt, font=\scriptsize] at (axis cs:72.10,2) {72.10$_{\textcolor{red}{(+1.00)}}$};

\addplot[
    draw=black,                           
    fill=myblue!75,                        
    line width=1.0pt,
    bar shift=0pt
] coordinates {
    (71.77,3)   
};
\node[anchor=west, xshift=2pt, yshift=-1pt, font=\scriptsize] at (axis cs:71.77,3) {71.77$_{\textcolor{red}{(+0.67)}}$};

\addplot[
    draw=black,                           
    fill=myblue,                        
    line width=1.0pt,
    bar shift=0pt
] coordinates {
    (72.54,4)  
};
\node[anchor=west, xshift=2pt, yshift=-1pt, font=\scriptsize] at (axis cs:72.54,4) {72.54$_{\textcolor{red}{(+1.44)}}$};

\end{axis}
\end{tikzpicture}

%% file: figures/abl_step_by_step_w4a4.tex
\begin{tikzpicture}
\begin{axis}[
    xbar,                                  %
    width=7cm,
    height=3.5cm,
    xmin=52.0, xmax=64.0,
    xtick={46,48,50,52,54,56,58,60,62,64,66,68}, %
    minor x tick num=3,
    y dir=reverse,                         %
    ytick={1,2,3,4},
    yticklabels={Naive,+SRL,+PBA,\textbf{Ours}},
    bar width=8pt,
    enlarge y limits=0.3,
    axis line style={thick}, %
    tick style={thick}, %
    ymajorgrids=false, %
    yminorgrids=false, %
    xmajorgrids=true, %
    xminorgrids=true, %
    major grid style={gray!45,dashed}, %
    minor grid style={gray!45,dotted}, %
]

\addplot[
    draw=black,                           
    fill=myred!25,                        
    line width=1.0pt,
    bar shift=0pt
] coordinates {
    (53.76,1)   
};
\node[anchor=west, xshift=2pt, yshift=0pt, font=\scriptsize] at (axis cs:53.76,1) {53.76};

\addplot[
    draw=black,                           
    fill=myred!50,                        
    line width=1.0pt,
    bar shift=0pt
] coordinates {
    (57.12,2)
};
\node[anchor=west, xshift=2pt, yshift=-1pt, font=\scriptsize] at (axis cs:57.12,2) {57.12$_{\textcolor{red}{(+3.36)}}$};

\addplot[
    draw=black,                           
    fill=myred!75,                        
    line width=1.0pt,
    bar shift=0pt
] coordinates {
    (56.78,3)   
};
\node[anchor=west, xshift=2pt, yshift=-1pt, font=\scriptsize] at (axis cs:56.78,3) {56.78$_{\textcolor{red}{(+3.02)}}$};

\addplot[
    draw=black,                           
    fill=myred,                        
    line width=1.0pt,
    bar shift=0pt
] coordinates {
    (59.83,4)  
};
\node[anchor=west, xshift=2pt, yshift=-1pt, font=\scriptsize] at (axis cs:59.83,4) {59.83$_{\textcolor{red}{(+6.07)}}$};

\end{axis}
\end{tikzpicture}

%% file: tables/abl_metric.tex
\begin{tabular}{
    >{\centering\arraybackslash}m{1.5cm}
    >{\centering\arraybackslash}m{1.1cm}
    >{\centering\arraybackslash}m{1.4cm}
}
\toprule
\textbf{Metric} & \textbf{Prior} & \textbf{Avg.} ($\uparrow$) \\
\midrule
Random & - & 71.88 \\
$\lVert\mathbf{G}_w\rVert_{\ell_1}$ & Lower & 72.10 \\
$\lVert\mathbf{G}_w\rVert_{\ell_1}$ & Higher & 72.19 \\
$\vec{\tau}$ & Lower & 72.03 \\
\bg $\vec{\tau}$ (Ours) & \bg Higher & \bg \textbf{72.54} \\
\bottomrule
\end{tabular}

%% file: tables/abl_scheduler.tex
\begin{tabular}{
    >{\centering\arraybackslash}m{2.4cm}
    >{\centering\arraybackslash}m{1.4cm}
}
\toprule
\textbf{Scheduler} & \textbf{Avg.} ($\uparrow$) \\
\midrule
Cosine  & 71.34 \\
Linear  & 71.00 \\
\bg Log. (Ours)  & \bg \textbf{72.54} \\
\bottomrule
\end{tabular}

%% file: tables/abl_block_granularity.tex
\begin{tabular}{
    >{\centering\arraybackslash}m{2.4cm}
    >{\centering\arraybackslash}m{1.4cm}
}
\toprule
\textbf{Blk. Granu.} & \textbf{Avg.} ($\uparrow$) \\
\midrule
$0.5 \times \theta_\text{DecLayer}$  & 72.15 \\
$\ \ 2 \times \theta_\text{DecLayer}$  & 70.28 \\
\bg $\ \ 1 \times \theta_\text{DecLayer}$  & \bg \textbf{72.54} \\
\bottomrule
\end{tabular}

%% file: tables/abl_labeler.tex
\begin{tabular}{
    >{\centering\arraybackslash}m{1.7cm}
    >{\centering\arraybackslash}m{1.2cm}
    >{\centering\arraybackslash}m{1.2cm}
}
\toprule
\textbf{Labeler} & \textbf{W4A8}  & \textbf{W4A4} \\
\midrule
GPT-4 & 71.77 & 56.78 \\
\bg Self (Ours) & \bg \textbf{72.54} & \bg \textbf{59.83} \\
Homo. 8B$^\ddag$ & 72.35 & 59.04 \\
\bottomrule
\end{tabular}

%% file: docs/conclusion.tex
\section{Conclusion}\label{sec:conclusion}
In this paper, we present \method, a weight-activation QAT framework tailored for low-bit quantization of small-scale MLLMs. 
To mitigate data mismatch under limited fine-tuning budgets, we introduced an SRL scheme that re-annotates open-source MM datasets using the target model, yielding supervision that more closely aligns with the model's distribution. To address the coupled degradation of MM understanding and language modeling, we proposed a PBA mechanism that prioritizes recovering MM understanding first, guided by a vision-sensitivity factor $\Vec{\tau}$ for identifying MM-critical layers. Extensive evaluations on diverse MM benchmarks and models demonstrate that \method achieves near-lossless accuracy under W4A8 and remains robust where PTQ methods collapse at W4A4. We expect \method to facilitate hardware-friendly deployment of MLLMs on resource-constrained edge platforms.

%% file: docs/appendix.tex
\appendix
\setcounter{equation}{0}
\setcounter{figure}{0}
\setcounter{table}{0}
\renewcommand{\theequation}{A\arabic{equation}}
\renewcommand{\thefigure}{A\arabic{figure}}
\renewcommand{\thetable}{A\arabic{table}}
\renewcommand{\theHequation}{appendix.\arabic{equation}}
\renewcommand{\theHfigure}{appendix.\arabic{figure}}
\renewcommand{\theHtable}{appendix.\arabic{table}}

\newcommand{\CRfit}[1]{\begingroup\setbox0=\hbox{#1}\ifdim\wd0>\linewidth\resizebox{\linewidth}{!}{\usebox0}\else\usebox0\fi\endgroup}

\section*{Appendix Overview}
We provide additional details in the appendix, organized as follows
\begin{itemize}[leftmargin=*, noitemsep, label=$\triangleright$]
    \item Section~\ref{app:preliminary} introduces MLLM quantization preliminaries.
    \item Section~\ref{app:method} provides additional details of \method, including the overall training procedure and extended SRL examples.
    \item Section~\ref{app:more_impl} reports implementation details, covering baseline configurations, SRL data generation settings, and the reported training configuration for QAT.
    \item Section~\ref{app:experiments} supplements the main results with additional ablations, cross-family evaluation, and SRL distribution and robustness analyses.
    \item Section~\ref{app:hardware} presents the hardware architecture, measurement setup, and BF16/W8A8/W4A8/W4A4 efficiency comparisons.
    \item Section~\ref{app:limitation} discusses limitations and broader impacts.
\end{itemize}

\section{Preliminary}\label{app:preliminary}
\subsection{Background}\label{app:background}
Existing QAT solutions primarily focus on low-bit weight-only quantization~\citep{chen2025efficientqat,du2024bitdistiller,huang2024billm,guo2025speedq}. 
However, weight-only inference typically relies on floating-point cores to process activations, incurring substantial power and area overhead on edge AI chips~\citep{van2023fp8-vs-int8,li2024llm-hardware-accelerator}. 
This practical constraint motivates us to pursue high-performance weight-activation quantization, and our work targets weight-activation QAT for low-bit edge MLLMs.

SmoothQuant-style methods~\citep{xiao2023smoothquant} achieve strong accuracy at W8A8, but pushing to lower-bit settings often depends on rotation-based quantization~\citep{ashkboos2024quarot,liu2024spinquant,hu2025ostquant,sun2024flatquant}. 
Although the overhead can be modest on GPUs~\citep{ashkboos2024quarot,sun2024flatquant}, such online rotations typically require additional hardware support on edge chips, increasing chip area and compute cost~\citep{kim2026isscc,liu2026isscc,dong2026isscc}. 
In contrast, \method introduces no extra runtime operators during inference beyond standard quantized linear layers.

\subsection{Quantization}
Quantization reduces the precision of weights (and/or activations) to save memory and accelerate computation. 
For a weight tensor $\mathbf{W}$, standard uniform $b$-bit quantization and de-quantization are formulated as
\begin{align}
    \mathbf{W}_\text{int} = \operatorname{clip}\left( \left\lfloor \frac{\mathbf{W}}{s} \right\rceil + z,\ q_{\min},\ q_{\max} \right),\quad
    \hat{\mathbf{W}} = \left( \mathbf{W}_\text{int} - z \right)\times s,
\end{align}
where $\mathbf{W}_\text{int}$ and $\hat{\mathbf{W}}$ denote the quantized integer weights and the de-quantized full-precision weights, respectively. 
Here, $s$ and $z$ are the scale and zero-point, and $\operatorname{clip}(\cdot,\cdot,\cdot)$ bounds integer values to $[q_{\min},q_{\max}]$. 
In this paper, we adopt static \textit{per-channel} quantization for weights and dynamic \textit{per-token} quantization for activations, following common practice in the community~\citep{liu2023llm,xiao2023smoothquant}. We use signed symmetric quantization with $z=0$ throughout; $q_{\min}$ and $q_{\max}$ denote the signed integer bounds of the quantizer.

\subsection{Training-based Quantization}
Training-based quantization introduces learnable parameters~\citep{xie2024qslaw,liu2023llm,liu2024spinquant} or functions~\citep{nagel2020adaround,li2021brecq} to recover quantized model accuracy in a block-wise~\citep{li2021brecq,shao2023omniquant} or network-wise manner~\citep{liu2023llm,du2024bitdistiller}. 
In this paper, \method performs network-level QAT by learning the quantization scales $s$ together with the weight parameters $\mathbf{W}$. 
To make the rounding operator $\lfloor\cdot\rceil$ differentiable, we adopt the straight-through estimator (STE)~\citep{bengio2013ste}. 
Following prior works~\citep{liu2023llm,du2024bitdistiller,li2026gear}, we employ self-distillation, where the full-precision model serves as the teacher to guide the quantized student. 
We also incorporate supervision from the training response labels, as in~\citep{guo2025speedq}. 
Specifically, our objective is a weighted sum of forward KL, reverse KL, and cross-entropy losses:
\begin{equation}\label{eq:loss_func}
    \mathcal{L}_\text{QAT} = \alpha \mathcal{L}_\text{KL}(P_s||P_t) + (1-\alpha) \mathcal{L}_\text{KL}(P_t||P_s) + \beta \mathcal{L}_\text{CE}(Y,P_s),
\end{equation}
where $\alpha$ is estimated at the start of training as the average token probability over a randomly sampled mini-batch~\citep{guo2025speedq,du2024bitdistiller}. 
We empirically set $\beta$ to match the scale of the other loss terms. 
Here, $P_t$ and $P_s$ are teacher and student token-probability distributions obtained by applying softmax to their logits. For \method, $Y=Y_{\mathrm{SRL}}$ denotes the self-relabeled response targets; labeler ablations replace these targets with the corresponding label set. More discussion is provided in Appendix~\ref{app:abl_loss_func}.

\section{\method}\label{app:method}

\subsection{Training Procedures}
Algorithm~\ref{alg:pseudocode} outlines the training procedure of \method. Starting from the re-labeled dataset $\mathcal{D}_{\text{SRL}}$, the full-precision teacher $\mathcal{M}_{\text{FP}}$, and the quantized student $\mathcal{M}_{\text{Q}}$, the model is optimized over $S$ steps with a predefined activation schedule $\mathbb{T}$ and block budget $\mathbb{B}$. At scheduled steps, \method performs teacher-student forward propagation, computes the QAT loss, updates the quantized parameters, and progressively activates more quantized blocks based on the estimated sensitivity. At all other steps, the student model is updated with standard QAT.

\input{figures/pseudocode}

\subsection{Examples of SRL}
\begin{figure}[t]
    \centering
    \includegraphics[width=\linewidth]{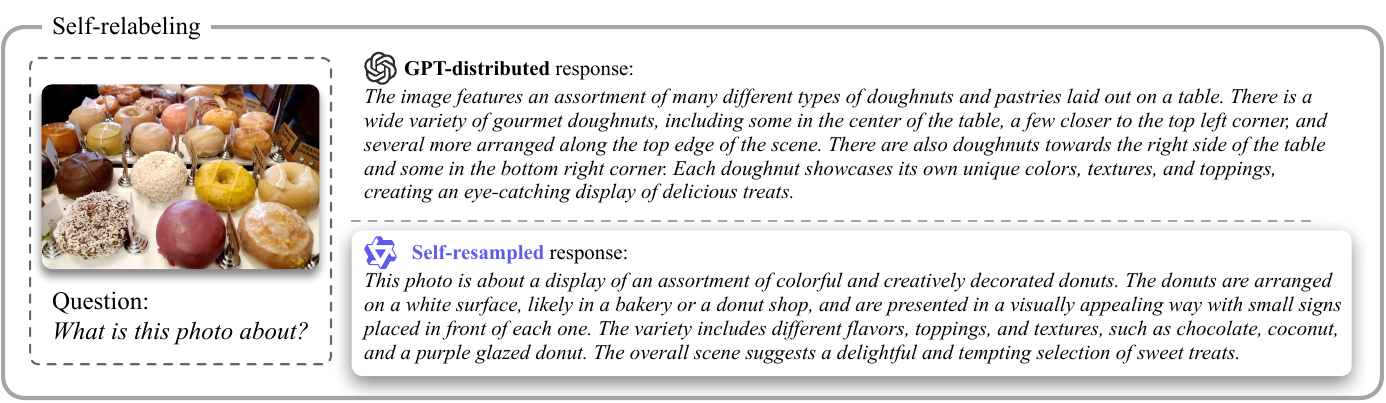}
    \caption{Extended example of resampled datasets in Section~\ref{sec:method}.}
    \label{fig:srl_examples}
\end{figure}
We present one full example from the COCO caption dataset~\citep{liu2024llava1.5} in Figure~\ref{fig:srl_examples}. It is evident that the two responses exhibit distinct stylistic characteristics, suggesting a potential distribution gap between them. Specifically, GPT-style responses tend to be more literal and spatially grounded, emphasizing object presence and layout with explicit positional phrases such as in the center of the table, closer to the top left corner, and along the top edge. In contrast, Qwen-style responses are more holistic, semantically enriched, and interpretive, focusing more on scene-level summarization, attributes, and visual appeal, as reflected in expressions such as colorful and creatively decorated donuts, different flavors, toppings, and textures, and a delightful and tempting selection of sweet treats. This discrepancy indicates that, even when describing the same image, different caption sources may encode substantially different linguistic preferences and descriptive priors. Such stylistic inconsistency can further introduce additional challenges for model adaptation and alignment, especially when the training objective implicitly assumes a unified target distribution.

\section{Implementation Details}\label{app:more_impl}
\subsection{Baselines}
\textbf{SmoothQuant~\citep{xiao2023smoothquant}.}
SmoothQuant alleviates activation outliers via an equivalent per-channel scaling transformation. It uses dynamic per-token activation quantization together with static per-channel weight quantization.

\textbf{MBQ~\citep{li2025mbq}.}
MBQ extends SmoothQuant by introducing modality-balanced scaling factors derived from gradient information. Its quantized inference configuration is kept the same as that of SmoothQuant.

\textbf{MQuant~\citep{yu2025mquant}.}
MQuant addresses the distributional discrepancy between visual and textual tokens, as well as the extreme outliers introduced by Hadamard-based transformations. It adopts a fully static quantization scheme for efficient deployment.

\textbf{LLM-QAT~\citep{liu2023llm}.}
LLM-QAT pioneers QAT framework for language transformers, with a primary focus on data-free distillation. Since its original data-free design is not directly applicable to MLLMs, we reproduce LLM-QAT using the same training data, budget, and optimization settings as \method. For the W4A4 backbones marked with $\dag$ in Table~\ref{tab:exp_main}, we retain 8-bit \texttt{down\_proj} activations for LLM-QAT to avoid severe degradation. The PTQ baselines are reproduced with their official implementations and the configurations reported in their respective papers.

\subsection{Data Generation}
We use the vLLM engine~\citep{kwon2023vllm} to generate the dataset on a single 8$\times$H200 node. 
Following the practice of LLM-QAT~\citep{liu2023llm}, we generate SRL responses via stochastic sampling rather than greedy decoding, as sampling produces more diverse outputs and yields better performance across tasks. 
We follow the default configuration files for sampling hyperparameters. 
For \texttt{Qwen3-VL-2B-Instruct}, we set $\mathtt{top\_p}=0.8$, $\mathtt{top\_k}=20$, $\mathtt{temperature}=0.7$, and $\mathtt{repetition\_penalty}=1.0$ according to the released $\mathtt{generation\_config.json}$. 
Additionally, we set $\mathtt{max\_token\_len}=512$ to avoid excessively long responses. 
Generating SRL labels for \texttt{LLaVA-680K}~\citep{liu2024llava1.5} takes $\sim$2.5 GPU days for 2B-scale models.

\subsection{Training Configuration}
We implement \method in \texttt{PyTorch} and train quantized models with \texttt{DeepSpeed}. Each experiment is conducted on a single 8$\times$H200 node. 
Unless otherwise specified, we fine-tune quantized models for 5 epochs in BF16, enabling gradient checkpoints and using gradient accumulation with 2 steps. 
We set the per-device batch size to 32 and the maximum sequence length to 8192 tokens. 
We use the \texttt{AdamW} optimizer with a constant learning rate, zero weight decay, and a maximum gradient norm of 1.0. 
During QAT, we freeze the vision transformer and train the remaining modules together with the quantization scales, using module-specific learning rates. 
We optimize a hybrid objective that combines forward KL, reverse KL, and supervised fine-tuning losses, as defined in Eq.~\ref{eq:loss_func}. The QAT takes $\sim$2 GPU days for 2B-scale models.

\section{More Experiments}\label{app:experiments}
\subsection{Ablation on SmoothQuant initialization.}
\method can be applied on top of SmoothQuant~\citep{xiao2023smoothquant} initialization. As shown in Table~\ref{tab:abl_smoothq}, combining SmoothQuant with W4A4 quantization further improves accuracy. In contrast, SmoothQuant can slightly degrade performance under W4A8 settings. These results suggest that SmoothQuant serves as an effective initialization for extreme low-bit (W4A4) QAT, consistent with observations in LLM-QAT~\citep{liu2023llm}.
\begin{table}[h]
    \centering
    \caption{\method \textit{w/} or \textit{w/o} SmoothQuant.}
    \resizebox{0.35\linewidth}{!}{\input{tables/abl_smoothq}}
    \label{tab:abl_smoothq}
\end{table}

\subsection{Ablation Study on Loss Functions}~\label{app:abl_loss_func}
We present performance under different supervision objectives for W4A8 \texttt{Qwen3-VL-2B-Instruct} QAT in Table~\ref{tab:abl_loss_func}. 
Using a single KL term yields limited gains, where forward KL achieves 71.76\% on average accuracy while reversed KL performs worse, suggesting that either direction alone provides insufficient guidance for stable recovery. Combining both terms via CAKLD~\citep{du2024bitdistiller} improves the average accuracy to 71.94\%, indicating that balancing the two divergences is beneficial. The balance factor $\alpha$ is estimated at the start of training as described above.
Building on CAKLD, our multi-objective formulation further incorporates a cross-entropy term following~\citep{guo2025speedq}, achieving the best performance. Overall, the results validate that jointly leveraging bidirectional distillation and label supervision provides the most effective training signal for low-bit QAT.
\begin{table}[h]
    \centering
    \caption{QAT performance under different supervisions.}
    \resizebox{0.7\linewidth}{!}{\input{tables/abl_loss_func}}
    \label{tab:abl_loss_func}
\end{table}

\subsection{Generalization Beyond the Qwen-VL Family}\label{app:cr_generalization}
We evaluate InternVL3.5-1B and LLaVA-OV-0.5B at W4A8 using the same seven-benchmark average as in the main experiments (Table~\ref{tab:cr_generalization}). \method achieves 63.93 and 52.29, respectively, outperforming MBQ and LLM-QAT on both models. These experiments extend the evaluated architectures beyond Qwen-VL.

\begin{table}[htbp]
\centering\small
\caption{Average accuracy over the seven benchmarks used in the main paper on two additional model families. Quantized results use W4A8.}
\label{tab:cr_generalization}
\CRfit{\input{tables/cr/generalization}}
\end{table}

Table~\ref{tab:cr_internvl_ablation} isolates the contributions of SRL and PBA on InternVL3.5-1B. Each component improves naive QAT, and their combination performs best.

\begin{table}[htbp]
\centering\small
\caption{Component ablation on InternVL3.5-1B at W4A8.}
\label{tab:cr_internvl_ablation}
\CRfit{\input{tables/cr/internvl_ablation}}
\end{table}

\subsection{Understanding SRL: Distribution, Style, and Label Transfer}\label{app:cr_srl_analysis}
We compare five label sets on the same randomly sampled 10K cases, pairing responses by prompt. The target is Qwen3-VL-2B and its self-labels serve as the reference. GPT and LV-0.5B/LV-7B are cross-family labelers; Qw-Homo-8B is a larger same-family labeler. The diagnostics in Table~\ref{tab:cr_srl_distribution} are defined as follows.
\begin{enumerate}[leftmargin=*,itemsep=2pt]
\item \textbf{NLL (nats/token).} We compute $-|y|^{-1}\sum_t\log p_{\mathrm{target}}(y_t\mid y_{<t},x)$. Lower values indicate higher likelihood under the target model.
\item \textbf{Vocabulary coverage.} The fraction of the reference's top-5K most frequent vocabulary items covered by assistant tokens in the label set.
\item \textbf{Normalized length distance.} The Wasserstein-1 distance between response token-count distributions, divided by the reference mean length.
\item \textbf{Format distance.} The $\ell_1$ distance between mean 13-dimensional binary format-feature vectors, including headings, bold text, bullets, lists, code blocks, LaTeX, and tables.
\item \textbf{Semantic distance.} The mean paired $1-\cos$ similarity between responses encoded with the frozen \texttt{gte-large-en-v1.5} sentence encoder~\citep{li2023towards}.
\end{enumerate}

\begin{table}[htbp]
\centering\small
\caption{Label-distribution diagnostics and QAT accuracy on Qwen3-VL-2B. Qw-Self-2B is the reference; LV labelers are cross-family and Qw-Homo-8B is a larger same-family labeler.}
\label{tab:cr_srl_distribution}
\CRfit{\input{tables/cr/srl_distribution}}
\end{table}

Response style is a component of the output distribution, rather than an alternative to distributional alignment. Vocabulary, length, and format characterize surface form, while embedding similarity characterizes semantics; these diagnostics are complementary views and are not an additive decomposition of NLL. Matching surface form alone does not explain the full gain: the same-family 8B labeler reaches reference vocabulary coverage (1.000), but its NLL remains 0.906 versus 0.474 for self-labels and its QAT accuracy remains lower.

At W4A4, the NLL ranking agrees exactly with the inverse accuracy ranking. At W4A8, GPT versus LV-7B is an exception (their NLL differs by 0.009). Moreover, LV-0.5B is closer to the reference than LV-7B on all four auxiliary diagnostics, but has higher NLL and lower QAT accuracy at both precisions. These results support target-distribution proximity as an explanation of SRL's gain beyond a single stylistic property.

\paragraph{Label transferability.}
Self-labels give the best W4A4 accuracy (59.83), followed by same-family 8B labels (59.04), and then cross-family labels (57.34, 56.78, and 56.31). Same-family reuse is therefore possible with an accuracy penalty, while cross-family reuse does not preserve the full benefit of SRL. We recommend per-target-model relabeling where its generation cost is acceptable.

\subsection{Hallucination and Bias Evaluation}\label{app:cr_robustness}
We evaluate Qwen2-VL-2B on HallusionBench~\citep{guan2024hallusionbench}, using DeepSeek-V4-Flash as the judge, and on VLMsAreBiased~\citep{vlmsarebiased}. Table~\ref{tab:cr_robustness} compares \method with and without SRL at W4A8 and W4A4. SRL improves the reported scores at both precisions, while the quantized models remain below the BF16 target. Thus, these evaluations do not show a deterioration from SRL on the two tested benchmarks. They do not rule out inherited hallucinations, biases, or failure modes~\citep{li2026token} outside these evaluations.

\begin{table}[htbp]
\centering\small
\caption{Hallucination and bias evaluation on Qwen2-VL-2B. HallusionBench uses DeepSeek-V4-Flash as the judge. The table reports the scores for each setting.}
\label{tab:cr_robustness}
\CRfit{\input{tables/cr/srl_robustness}}
\end{table}

\clearpage
\section{Hardware Efficiency on Edge Accelerator}\label{app:hardware}
\begin{figure}[t]
    \centering
    \includegraphics[width=0.7\linewidth]{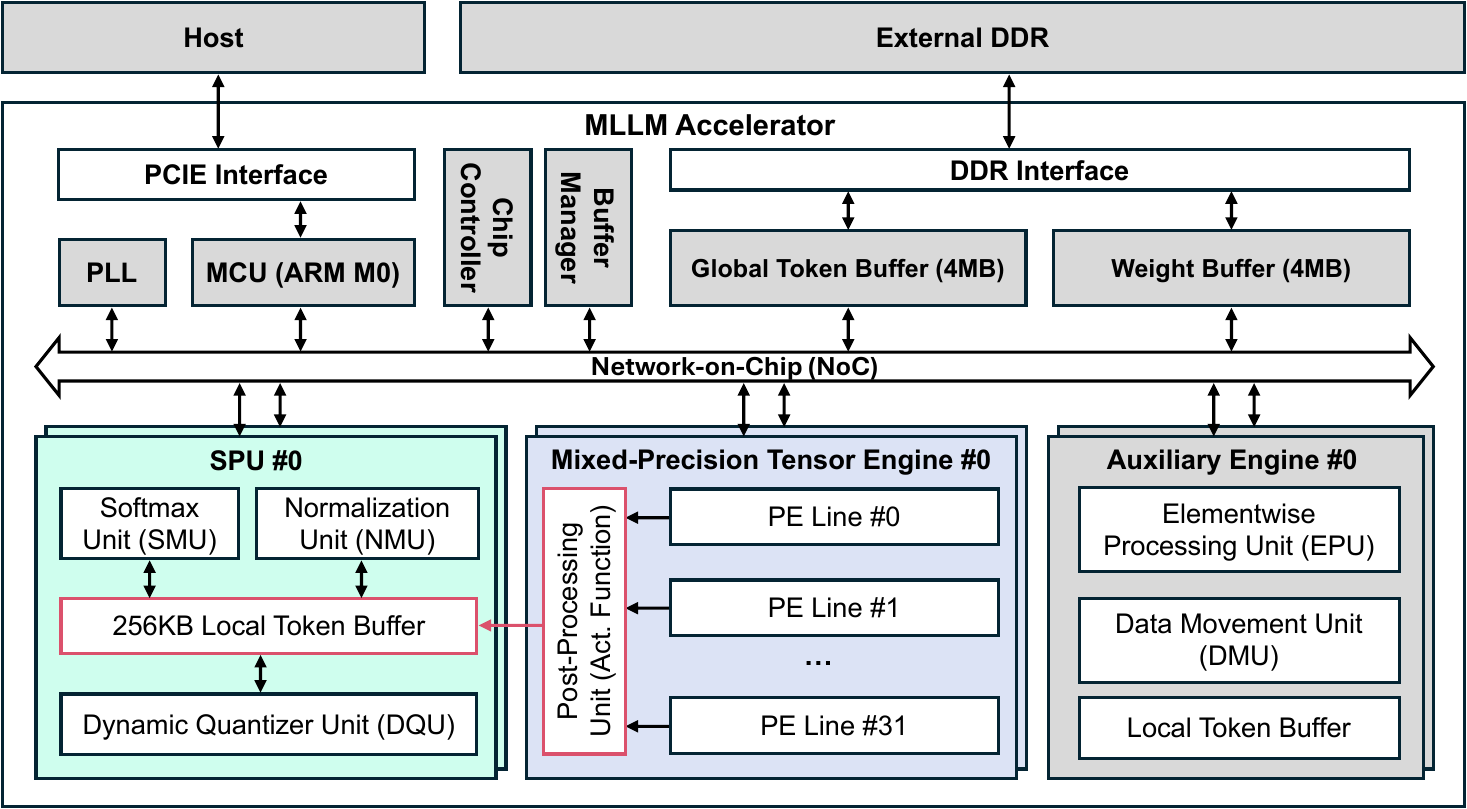}
    \caption{System-level architecture of our benchmark edge MLLM accelerator.}
    \label{fig:accelerator}
\end{figure}
\subsection{Hardware Architecture}\label{app:hardware_arch}
We evaluate an edge MLLM accelerator using measured test-chip behavior and a cycle-accurate accelerator model, as detailed in Section~\ref{app:cr_measurement}. Figure~\ref{fig:accelerator} shows its architecture. The accelerator is equipped with two Mixed-Precision Tensor Engines (MPTEs), each of which comprises 1K MACs organized into 32 PE lines under an output-stationary dataflow, delivering 2.46 TOPS at INT8 precision. The MPTEs are able to support BF16, INT8 and INT4 precision for both weights and activations. Besides, two Special Processing Units (SPUs) support BF16 softmax, normalization, and dynamic quantization. Furthermore, output feature map tiles from the MPTEs can be bypassed directly to the SPU's local token buffer to facilitate layer fusion, effectively overlapping the dynamic quantization latency. Two Auxiliary Engines handle element-wise operations, such as RoPE, and various data movement tasks like reshaping and transposing. The accelerator also features a 4\,MB global token buffer and a 4\,MB weight buffer, connected via a DDR4 interface that delivers 38.4 GB/s of external memory bandwidth. The compute engines and on-chip buffers are interconnected via a network-on-chip (NoC).

\subsection{System Setup and Measurement Methodology}\label{app:cr_measurement}
\begin{figure}[htbp]
\centering
\includegraphics[width=0.95\linewidth]{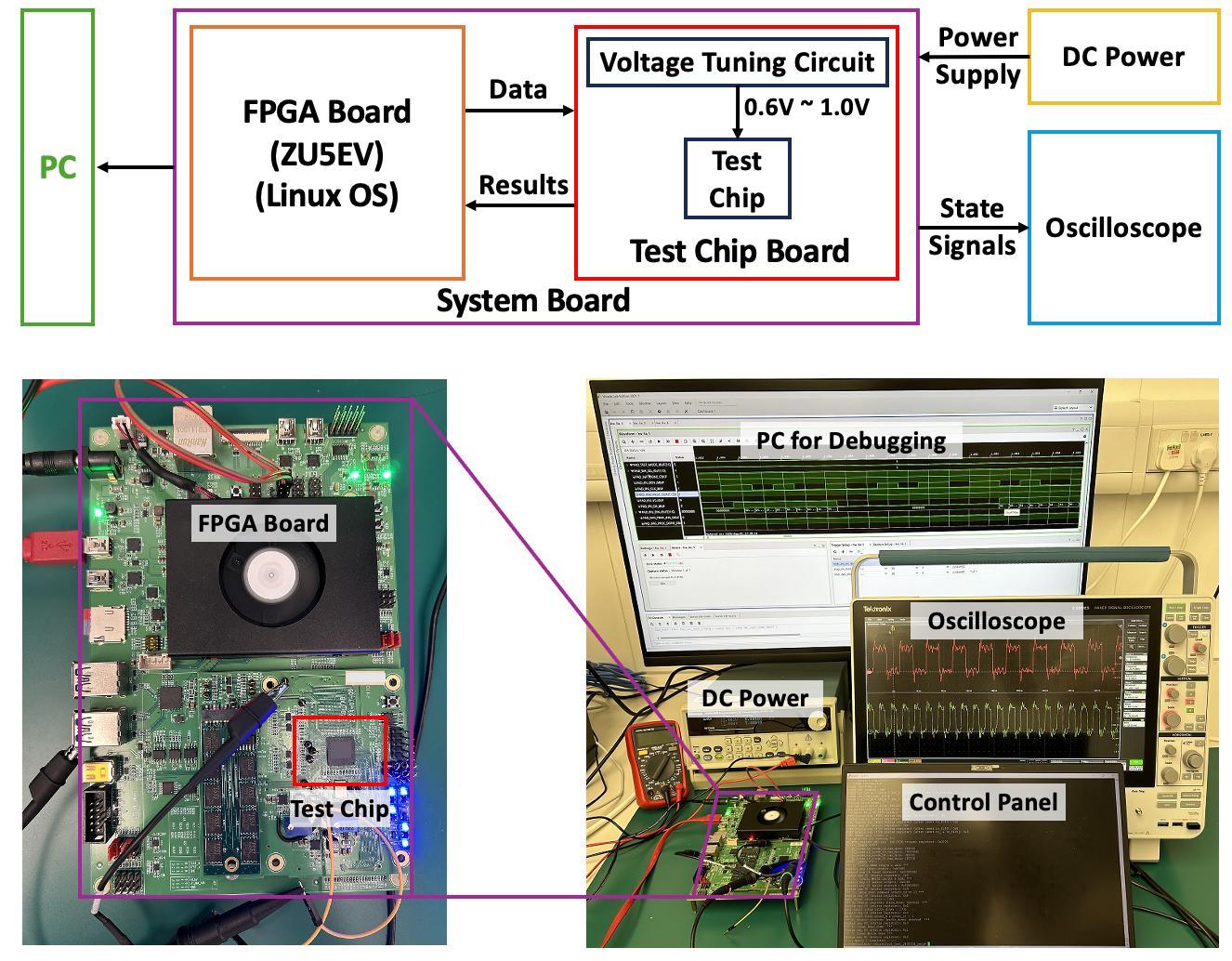}
\caption{Test environment. A Linux-based ZU5EV FPGA board supplies data and instructions to the test-chip board and receives results. The PC supports debugging and control, the DC supply powers the system board, and the oscilloscope observes state signals. The voltage tuning circuit supports 0.6--1.0\,V; this range is not a statement of the operating voltage used in every experiment.}
\label{fig:cr_hardware_setup}
\end{figure}
\paragraph{Hardware and memory.}
The test chip is a silicon-proven ASIC fabricated in TSMC 22\,nm CMOS. The FPGA is the interface/control board, rather than the inference accelerator. The system includes 8\,GB off-chip DDR4, a 4\,MB global token buffer for intermediate features, and a 4\,MB weight buffer for the current weight tile. The off-chip capacity accommodates the evaluated model weights when quantized to 8 bits or below. These are provisioned capacities, not measured peak memory usage.

\paragraph{Connections and workflow.}
The DC supply powers the system board; the PC monitors it and the oscilloscope observes its state signals. The voltage tuning circuit supplies the test chip. The FPGA sends data to the test-chip board and receives results. A customized MLIR-based compiler produces test data and instructions. The FPGA forwards them to the chip, and latency is monitored via the PC waveform.

\paragraph{Measured and simulated components.}
On-chip energy is obtained from measured latency and average power,
\begin{equation}
E_{\mathrm{chip}}=t_{\mathrm{chip}}\,\overline{P}_{\mathrm{chip}}.
\end{equation}
Off-chip DRAM access latency and energy are simulated using DRAMPower. The end-to-end cycle-accurate accelerator model combines the on-chip and off-chip contributions. The reported boundary includes the accelerator and DRAM accesses, not the wall-plug consumption of the PC, FPGA control board, oscilloscope, and other test equipment.

\paragraph{Reported metrics.}
Throughput is calculated as $1000/t$, where $t$ is the reported latency in milliseconds. For the same token configuration, speedup is $t_{\mathrm{BF16}}/t_{\mathrm{quant}}$ and the energy-reduction ratio is $E_{\mathrm{BF16}}/E_{\mathrm{quant}}$. The tables distinguish prefill and decode and preserve the original latency/energy units.

\subsection{Edge Evaluation Results}\label{app:more_hardware_results}
Tables~\ref{tab:acceleration_qwen2_vl_2b}, \ref{tab:acceleration_qwen2_5_vl_4b}, and~\ref{tab:acceleration_qwen3_vl_2b} present comprehensive latency and energy measurements for \texttt{Qwen2-VL-2B-Instruct}, \texttt{Qwen2.5-VL-3B-Instruct} and \texttt{Qwen3-VL-2B-Instruct} on the edge MLLM accelerator. Across all three models, the prefill stage shows W4A8 achieving 1.48$\sim$1.89$\times$ speedup and 1.94$\sim$3.13$\times$ energy savings, while W4A4 further attains 2.11$\sim$3.64$\times$ speedup and 2.53$\sim$5.64$\times$ energy reduction. The decode stage exhibits consistent 3.74$\sim$3.98$\times$ speedup and 3.76$\sim$3.99$\times$ energy savings for both W4A8 and W4A4 due to memory-bound issue. Notably, the speedup ratio diminishes with longer visual token length since the ViT FLOPs proportion increases. As the ViT backbone typically remains in BF16 precision to preserve visual features, it becomes a bottleneck on edge accelerators where BF16 compute resources are insufficient due to area constraints. These results validate that \method generalizes effectively across model scales and delivers substantial efficiency gains for edge deployment.

\begin{table}[htbp]
    \centering
    \caption{Performance comparison of BF16, W4A8, and W4A4 for \texttt{Qwen2-VL-2B-Instruct} in both prefill and decode stages. ``\#V Tok.'' and ``\#T Tok.'' denote the numbers of vision tokens and text tokens, respectively; ``Lat.'' and ``Eng.'' denote latency (ms) and energy (mJ), respectively; ``Spd.'' and ``Sav.'' denote the speedup and energy saving relative to the BF16 baseline, respectively.}
    \resizebox{\linewidth}{!}{\input{tables/acceleration_qwen2_vl_2b}}
    \label{tab:acceleration_qwen2_vl_2b}
\end{table}

\begin{table}[htbp]
    \centering
    \caption{Performance comparison of BF16, W4A8, and W4A4 for \texttt{Qwen2.5-VL-3B-Instruct} in both prefill and decode stages.}
    \resizebox{\linewidth}{!}{\input{tables/acceleration_qwen2_5_vl_4b}}
    \label{tab:acceleration_qwen2_5_vl_4b}
\end{table}

\begin{table}[htbp]
    \centering
    \caption{Performance comparison of BF16, W4A8, and W4A4 for \texttt{Qwen3-VL-2B-Instruct} in both prefill and decode stages.}
    \resizebox{\linewidth}{!}{\input{tables/acceleration_qwen3_vl_2b}}
    \label{tab:acceleration_qwen3_vl_2b}
\end{table}

\subsection{W8A8 Deployment and Precision Trade-offs}\label{app:cr_w8a8}
Tables~\ref{tab:cr_w8a8_qwen3}--\ref{tab:cr_w8a8_qwen25} add W8A8 deployment results for all three backbones. Table~\ref{tab:cr_deployment_ranges} summarizes the precision-dependent gains alongside the existing W4A8/W4A4 measurements. W8A8 gives about $1.00$--$1.05\times$ prefill speedup and $1.42$--$1.79\times$ energy reduction. Decode speedup is $1.95$--$2.00\times$, with $1.99$--$2.00\times$ energy reduction.

Moving from W8A8 to W4A8 halves weight traffic and primarily improves memory-bound decode. Moving from W4A8 to W4A4 additionally lowers activation precision and benefits prefill. Relative to W8A8, the supported W4A8 and W4A4 configurations offer approximately $2\times$ and $4\times$ peak compute throughput, respectively; these hardware peak ratios are not end-to-end measured speedups. Off-chip data movement can be energy-intensive, motivating the reduction in weight traffic.

\begin{table}[htbp]
\centering\small
\caption{BF16 / W8A8 deployment results for Qwen3-VL-2B. Throughput is in $s^{-1}$, latency in ms, and energy in mJ. Ratios are relative to BF16 under the same token configuration.}
\label{tab:cr_w8a8_qwen3}
\CRfit{\input{tables/cr/w8a8_qwen3}}
\end{table}

\begin{table}[htbp]
\centering\small
\caption{BF16 / W8A8 deployment results for Qwen2-VL-2B. Throughput is in $s^{-1}$, latency in ms, and energy in mJ. Ratios are relative to BF16 under the same token configuration.}
\label{tab:cr_w8a8_qwen2}
\CRfit{\input{tables/cr/w8a8_qwen2}}
\end{table}

\begin{table}[htbp]
\centering\small
\caption{BF16 / W8A8 deployment results for Qwen2.5-VL-3B. Throughput is in $s^{-1}$, latency in ms, and energy in mJ. Ratios are relative to BF16 under the same token configuration.}
\label{tab:cr_w8a8_qwen25}
\CRfit{\input{tables/cr/w8a8_qwen25}}
\end{table}

\begin{table}[htbp]
\centering\small
\caption{Latency speedup and energy-reduction ranges relative to BF16 across the four token configurations.}
\label{tab:cr_deployment_ranges}
\CRfit{\input{tables/cr/deployment_ranges}}
\end{table}

\section{Limitations and Broader Impacts}\label{app:limitation}

\paragraph{Limitations.}
Our experiments focus on compact MLLMs, with evaluated models up to 8B parameters. Extending \method to larger models and a broader range of multimodal architectures remains an important direction for future work.

\paragraph{Broader impacts.}
By reducing inference memory requirements, latency, and energy consumption in the evaluated deployment setting, \method can facilitate multimodal intelligence on resource-constrained devices. More efficient local inference may broaden access to multimodal applications and reduce reliance on remote
computing infrastructure.

%% file: figures/pseudocode.tex
\begin{algorithm}[h]
    \caption{\method \texttt{trainer}}
    \label{alg:pseudocode}
    \textbf{Preparation:} re-labeled finetuning datasets $\mathcal{D}_\text{SRL}=(X,\ Y_\text{SRL})$, full precision teacher model $\mathcal{M}_\text{FP}$, quantized student model $\mathcal{M}_\text{Q}$ with $L$ layers, and training steps $S$. \\
    \textbf{PBA toolkit:} activation timing $\mathbb{T}=\left[ S\times0\%,\ S\times30\%,\ S\times40\%,\ S\times50\% \right]$, and quantized block budget $\mathbb{B} = \left[ L\times50\%,\ L\times75\%, L\times88\%, L\times100\% \right]$.
    \begin{algorithmic}[0] %
    \For{$s = 0,1,2,\dots,S$}
        \If{$s \in \mathbb{T}$}
            \State $\mathtt{logits}_\text{FP} \gets \mathtt{forward} \left(\mathcal{M}_\text{FP}, \mathcal{D}_\text{SRL}^s \right)$; \Comment{compute teacher logits}
            \State $\mathtt{logits}_\text{Q} \gets \mathtt{forward} \left(\mathcal{M}_\text{Q}, \mathcal{D}_\text{SRL}^s \right)$; \Comment{compute student logits}
            \State $\mathcal{L}_\text{QAT}\gets\mathtt{loss}(\mathtt{logits}_\text{FP},\ \mathtt{logits}_\text{Q},\ Y_\text{SRL})$; \Comment{compute loss in Eq.~\ref{eq:loss_func}}
            \State $\vec{\tau},\ \mathbf{G} \gets \mathtt{backward}\left( \mathcal{M}_\text{Q}, \mathcal{L}_\text{QAT} \right)$; \Comment{cache $\vec{\tau}$ in Eq.~\ref{eq:vision_sensitivity}}
            \State $\widetilde{\mathbf{W}},\ \widetilde{\mathbf{S}} \gets \mathtt{optimize}(\mathbf{W},\ \mathbf{S};\ \mathbf{G})$; \Comment{update weight and scale}
            \State $\mathcal{M}_\text{Q} \gets \mathtt{block\_quant\_act}\left( \mathcal{M}_\text{Q};\ \mathtt{argsort}_{\downarrow}\left(\vec{\tau}\right),\ \mathbb{B}\right)$; \Comment{PBA mechanism}
        \Else
            \State $\mathtt{standard\_qat\_update}\left(\mathcal{M}_\text{Q};\ \mathcal{M}_\text{FP},\ \mathcal{D}_\text{SRL} \right)$;
        \EndIf
    \EndFor
    \State \textbf{Return} $\mathcal{M}_{\text{Q}}$
    \end{algorithmic}
\end{algorithm}

%% file: tables/abl_smoothq.tex
\begin{tabular}{
    >{\centering\arraybackslash}m{1.3cm}
    >{\centering\arraybackslash}m{1.4cm}
    >{\centering\arraybackslash}m{1.4cm}
    >{\centering\arraybackslash}m{1.4cm}
}
\toprule
\textbf{\#BW} & \textbf{SQ Init} & \textbf{Avg.} ($\uparrow$) & \textbf{MME} \\
\midrule
\multirow{2}{*}{W4A8}
& \ding{51} & 71.41 & 1956.58 \\
& \bg \ding{55} & \bg \textbf{72.54} & \bg \textbf{1962.04} \\
\midrule
\multirow{2}{*}{W4A4}
& \ding{55} & 50.87 & 1459.14 \\
& \bg \ding{51} & \bg \textbf{59.83} & \bg \textbf{1652.63} \\
\bottomrule
\end{tabular}

%% file: tables/abl_loss_func.tex
\begin{tabular}{
    >{\centering\arraybackslash}m{4.2cm}
    >{\centering\arraybackslash}m{4.5cm}
    >{\centering\arraybackslash}m{1.4cm}
}
\toprule
\textbf{Supervision} & \textbf{Formulation} & \textbf{Avg.} ($\uparrow$)\\
\midrule
KL Div. & $\mathcal{L}_\text{KL} = \mathcal{D}(P_t||P_s)$ & 71.76 \\
Reversed KL & $\mathcal{L}_\text{ReKL} = \mathcal{D}(P_s||P_t)$ & 71.41 \\
CAKLD~\cite{du2024bitdistiller} & $\mathcal{L}_\text{CAKLD} = \alpha \mathcal{L}_\text{ReKL} + (1-\alpha) \mathcal{L}_\text{KL}$  & 71.94 \\
\rowbg Multi-obj.~\cite{guo2025speedq} & $\mathcal{L}_\text{QAT} = \mathcal{L}_\text{CAKLD} + \beta \mathcal{L}_\text{CE}$  & \textbf{72.54} \\
\bottomrule
\end{tabular}

%% file: tables/cr/generalization.tex
\begin{tabular}{lrr}
\toprule
Method & InternVL3.5-1B & LLaVA-OV-0.5B \\
\midrule
BF16 & 69.92 & 57.95 \\
MBQ & 56.77 & 48.97 \\
LLM-QAT & 60.02 & 50.05 \\
ProQuant (ours) & 63.93 & 52.29 \\
\bottomrule
\end{tabular}

%% file: tables/cr/internvl_ablation.tex
\begin{tabular}{lr}
\toprule
Setting & InternVL3.5-1B Avg. \\
\midrule
Naive QAT & 61.74 \\
+ SRL & 62.98 (+1.24) \\
+ PBA & 63.01 (+1.27) \\
ProQuant & 63.93 (+2.19) \\
\bottomrule
\end{tabular}

%% file: tables/cr/srl_distribution.tex
\begin{tabular}{lrrrrr}
\toprule
Metric & GPT & LV-0.5B & LV-7B & Qw-Homo-8B & Qw-Self-2B \\
\midrule
NLL nats/token ($\downarrow$) & 1.459 & 1.504 & 1.450 & 0.906 & 0.474 \\
Vocab cov. top-5k ($\uparrow$) & 0.973 & 0.987 & 0.951 & 1.000 & 1 \\
Length W1, normalized ($\downarrow$) & 0.641 & 0.780 & 0.929 & 0.459 & 0 \\
Format L1 ($\downarrow$) & 2.831 & 2.861 & 3.268 & 1.213 & 0 \\
1 - paired cosine ($\downarrow$) & 0.178 & 0.172 & 0.265 & 0.067 & $\sim$0 \\
W4A8 Avg. Acc. ($\uparrow$) & 71.77 & 70.71 & 71.29 & 72.35 & 72.54 \\
W4A4 Avg. Acc. ($\uparrow$) & 56.78 & 56.31 & 57.34 & 59.04 & 59.83 \\
\bottomrule
\end{tabular}

%% file: tables/cr/srl_robustness.tex
\begin{tabular}{lrr}
\toprule
Setting & HallusionBench & VLMsAreBiased \\
\midrule
Full (BF16, target model) & 51.21 & 17.71 \\
ProQuant W4A8 & 50.37 & 17.24 \\
ProQuant w/o SRL W4A8 & 50.16 & 15.56 \\
ProQuant W4A4 & 43.84 & 14.58 \\
ProQuant w/o SRL W4A4 & 41.42 & 13.00 \\
\bottomrule
\end{tabular}

%% file: tables/acceleration_qwen2_vl_2b.tex
\begin{tabular}{
    >{\centering\arraybackslash}m{1.4cm}
    >{\centering\arraybackslash}m{1.5cm}
    >{\centering\arraybackslash}m{1.4cm}
    >{\centering\arraybackslash}m{1.5cm}
    >{\centering\arraybackslash}m{1.5cm}
    >{\centering\arraybackslash}m{1.5cm}
    >{\centering\arraybackslash}m{1.5cm}
    >{\centering\arraybackslash}m{1.5cm}
    >{\centering\arraybackslash}m{1.5cm}
    >{\centering\arraybackslash}m{1.5cm}
    >{\centering\arraybackslash}m{1.5cm}
    >{\centering\arraybackslash}m{1.5cm}
    >{\centering\arraybackslash}m{1.5cm}
}
\toprule
\multirow{2}{*}{\textbf{Stage}} & \multirow{2}{*}{\textbf{\#V Tok.}} & \multirow{2}{*}{\textbf{\#T Tok.}} 
& \multicolumn{2}{c}{\textbf{BF16}} 
& \multicolumn{4}{c}{\textbf{W4A8}} 
& \multicolumn{4}{c}{\textbf{W4A4}} \\
\cmidrule(lr){4-5}\cmidrule(lr){6-9}\cmidrule(lr){10-13}
& & & \textbf{Lat.} & \textbf{Eng.} & \textbf{Lat.} & \textbf{Spd.} & \textbf{Eng.} & \textbf{Sav.} & \textbf{Lat.} & \textbf{Spd.} & \textbf{Eng.} & \textbf{Sav.} \\
\midrule
\multirow{4}{*}{Prefill}
& \multirow{2}{*}{256}
& 256  & 877.02  & 2400.87 & 530.80  & \speedupbg 1.65$\times$ & 1020.90 & \savingbg 2.35$\times$ & 350.51  & \speedupbg 2.50$\times$ & 779.96  & \savingbg 3.08$\times$ \\
& 
& 1024 & 1977.79 & 4817.34 & 1144.89 & \speedupbg 1.73$\times$ & 1888.29 & \savingbg 2.55$\times$ & 632.23  & \speedupbg 3.13$\times$ & 1164.54 & \savingbg 4.14$\times$ \\
\cmidrule(lr){2-13}
& \multirow{2}{*}{1024}
& 256  & 2559.89 & 6045.12 & 1726.99 & \speedupbg 1.48$\times$ & 3116.07 & \savingbg 1.94$\times$ & 1214.34 & \speedupbg 2.11$\times$ & 2392.32 & \savingbg 2.53$\times$ \\
&
& 1024 & 3766.04 & 8654.54 & 2441.82 & \speedupbg 1.54$\times$ & 4069.73 & \savingbg 2.13$\times$ & 1522.48 & \speedupbg 2.47$\times$ & 2831.59 & \savingbg 3.06$\times$ \\
\midrule
\multirow{4}{*}{Decode}
& \multirow{2}{*}{256}
& 256  & 124.55 & 493.04 & 31.79 & \speedupbg 3.92$\times$ & 124.33 & \savingbg 3.97$\times$ & 31.40 & \speedupbg 3.97$\times$ & 123.95 & \savingbg 3.98$\times$ \\
&
& 1024 & 125.59 & 496.69 & 32.83 & \speedupbg 3.83$\times$ & 126.66 & \savingbg 3.92$\times$ & 31.85 & \speedupbg 3.94$\times$ & 125.71 & \savingbg 3.95$\times$ \\
\cmidrule(lr){2-13}
& \multirow{2}{*}{1024}
& 256  & 125.59 & 496.69 & 32.83 & \speedupbg 3.83$\times$ & 126.66 & \savingbg 3.92$\times$ & 31.85 & \speedupbg 3.94$\times$ & 125.71 & \savingbg 3.95$\times$ \\
&
& 1024 & 126.63 & 500.34 & 33.87 & \speedupbg 3.74$\times$ & 128.99 & \savingbg 3.88$\times$ & 32.30 & \speedupbg 3.92$\times$ & 127.46 & \savingbg 3.93$\times$ \\
\bottomrule
\end{tabular}

%% file: tables/acceleration_qwen2_5_vl_4b.tex
\begin{tabular}{
    >{\centering\arraybackslash}m{1.4cm}
    >{\centering\arraybackslash}m{1.5cm}
    >{\centering\arraybackslash}m{1.4cm}
    >{\centering\arraybackslash}m{1.5cm}
    >{\centering\arraybackslash}m{1.5cm}
    >{\centering\arraybackslash}m{1.5cm}
    >{\centering\arraybackslash}m{1.5cm}
    >{\centering\arraybackslash}m{1.5cm}
    >{\centering\arraybackslash}m{1.5cm}
    >{\centering\arraybackslash}m{1.5cm}
    >{\centering\arraybackslash}m{1.5cm}
    >{\centering\arraybackslash}m{1.5cm}
    >{\centering\arraybackslash}m{1.5cm}
}
\toprule
\multirow{2}{*}{\textbf{Stage}} & \multirow{2}{*}{\textbf{\#V Tok.}} & \multirow{2}{*}{\textbf{\#T Tok.}} 
& \multicolumn{2}{c}{\textbf{BF16}} 
& \multicolumn{4}{c}{\textbf{W4A8}} 
& \multicolumn{4}{c}{\textbf{W4A4}} \\
\cmidrule(lr){4-5}\cmidrule(lr){6-9}\cmidrule(lr){10-13}
& & & \textbf{Lat.} & \textbf{Eng.} & \textbf{Lat.} & \textbf{Spd.} & \textbf{Eng.} & \textbf{Sav.} & \textbf{Lat.} & \textbf{Spd.} & \textbf{Eng.} & \textbf{Sav.} \\
\midrule
\multirow{4}{*}{Prefill}
& \multirow{2}{*}{256}
& 256  & 1663.57 & 4802.11  & 882.53  & \speedupbg 1.89$\times$ & 1580.31 & \savingbg 3.04$\times$ & 507.40  & \speedupbg 3.28$\times$ & 1030.43 & \savingbg 4.66$\times$ \\
&
& 1024 & 3959.49 & 10486.84 & 2132.48 & \speedupbg 1.86$\times$ & 3347.68 & \savingbg 3.13$\times$ & 1088.50 & \speedupbg 3.64$\times$ & 1859.23 & \savingbg 5.64$\times$ \\
\cmidrule(lr){2-13}
& \multirow{2}{*}{1024}
& 256  & 4439.11 & 11419.90 & 2612.11 & \speedupbg 1.70$\times$ & 4280.74 & \savingbg 2.67$\times$ & 1568.12 & \speedupbg 2.83$\times$ & 2792.29 & \savingbg 4.09$\times$ \\
&
& 1024 & 6972.05 & 17529.33 & 4034.06 & \speedupbg 1.73$\times$ & 6226.26 & \savingbg 2.82$\times$ & 2193.81 & \speedupbg 3.18$\times$ & 3719.84 & \savingbg 4.71$\times$ \\
\midrule
\multirow{4}{*}{Decode}
& \multirow{2}{*}{256}
& 256  & 248.69 & 984.06 & 63.25 & \speedupbg 3.93$\times$ & 247.63 & \savingbg 3.97$\times$ & 62.45 & \speedupbg 3.98$\times$ & 246.85 & \savingbg 3.99$\times$ \\
&
& 1024 & 250.47 & 989.19 & 65.03 & \speedupbg 3.85$\times$ & 251.06 & \savingbg 3.94$\times$ & 63.03 & \speedupbg 3.97$\times$ & 249.11 & \savingbg 3.97$\times$ \\
\cmidrule(lr){2-13}
& \multirow{2}{*}{1024}
& 256  & 250.47 & 989.19 & 65.03 & \speedupbg 3.85$\times$ & 251.06 & \savingbg 3.94$\times$ & 63.03 & \speedupbg 3.97$\times$ & 249.11 & \savingbg 3.97$\times$ \\
&
& 1024 & 252.25 & 994.31 & 66.81 & \speedupbg 3.78$\times$ & 254.48 & \savingbg 3.91$\times$ & 63.61 & \speedupbg 3.97$\times$ & 251.37 & \savingbg 3.96$\times$ \\
\bottomrule
\end{tabular}

%% file: tables/acceleration_qwen3_vl_2b.tex
\begin{tabular}{
    >{\centering\arraybackslash}m{1.4cm}
    >{\centering\arraybackslash}m{1.5cm}
    >{\centering\arraybackslash}m{1.4cm}
    >{\centering\arraybackslash}m{1.5cm}
    >{\centering\arraybackslash}m{1.5cm}
    >{\centering\arraybackslash}m{1.5cm}
    >{\centering\arraybackslash}m{1.5cm}
    >{\centering\arraybackslash}m{1.5cm}
    >{\centering\arraybackslash}m{1.5cm}
    >{\centering\arraybackslash}m{1.5cm}
    >{\centering\arraybackslash}m{1.5cm}
    >{\centering\arraybackslash}m{1.5cm}
    >{\centering\arraybackslash}m{1.5cm}
}
\toprule
\multirow{2}{*}{\textbf{Stage}} & \multirow{2}{*}{\textbf{\#V Tok.}} & \multirow{2}{*}{\textbf{\#T Tok.}} 
& \multicolumn{2}{c}{\textbf{BF16}} 
& \multicolumn{4}{c}{\textbf{W4A8}} 
& \multicolumn{4}{c}{\textbf{W4A4}} \\
\cmidrule(lr){4-5}\cmidrule(lr){6-9}\cmidrule(lr){10-13}
& & & \textbf{Lat.} & \textbf{Eng.} & \textbf{Lat.} & \textbf{Spd.} & \textbf{Eng.} & \textbf{Sav.} & \textbf{Lat.} & \textbf{Spd.} & \textbf{Eng.} & \textbf{Sav.} \\
\midrule
\multirow{4}{*}{Prefill}
& \multirow{2}{*}{256}
& 256  & 869.62  & 2227.72 & 498.60  & \speedupbg 1.74$\times$ & 860.56  & \savingbg 2.59$\times$ & 300.42  & \speedupbg 2.89$\times$ & 569.67  & \savingbg 3.91$\times$ \\
&
& 1024 & 2090.36 & 4832.36 & 1190.86 & \speedupbg 1.76$\times$ & 1791.66 & \savingbg 2.70$\times$ & 612.83  & \speedupbg 3.41$\times$ & 1038.96 & \savingbg 4.65$\times$ \\
\cmidrule(lr){2-13}
& \multirow{2}{*}{1024}
& 256  & 2444.41 & 5570.82 & 1544.91 & \speedupbg 1.58$\times$ & 2530.13 & \savingbg 2.20$\times$ & 966.88  & \speedupbg 2.53$\times$ & 1777.43 & \savingbg 3.13$\times$ \\
&
& 1024 & 3798.92 & 8333.40 & 2370.94 & \speedupbg 1.60$\times$ & 3559.28 & \savingbg 2.34$\times$ & 1313.97 & \speedupbg 2.89$\times$ & 2231.03 & \savingbg 3.74$\times$ \\
\midrule
\multirow{4}{*}{Decode}
& \multirow{2}{*}{256}
& 256  & 140.46 & 556.20 & 35.89 & \speedupbg 3.91$\times$ & 141.56 & \savingbg 3.93$\times$ & 35.89 & \speedupbg 3.91$\times$ & 141.56 & \savingbg 3.93$\times$ \\
&
& 1024 & 144.00 & 570.20 & 37.66 & \speedupbg 3.82$\times$ & 148.57 & \savingbg 3.84$\times$ & 37.66 & \speedupbg 3.82$\times$ & 148.57 & \savingbg 3.84$\times$ \\
\cmidrule(lr){2-13}
& \multirow{2}{*}{1024}
& 256  & 144.00 & 570.20 & 37.66 & \speedupbg 3.82$\times$ & 148.57 & \savingbg 3.84$\times$ & 37.66 & \speedupbg 3.82$\times$ & 148.57 & \savingbg 3.84$\times$ \\
&
& 1024 & 147.53 & 584.20 & 39.44 & \speedupbg 3.74$\times$ & 155.57 & \savingbg 3.76$\times$ & 39.44 & \speedupbg 3.74$\times$ & 155.57 & \savingbg 3.76$\times$ \\
\bottomrule
\end{tabular}

%% file: tables/cr/w8a8_qwen3.tex
\begin{tabular}{lrrrrrrr}
\toprule
Stage & V tokens & T tokens & Throughput & Latency & Energy & Speedup & Energy ratio \\
\midrule
Prefill & 256 & 256 & 1.15 / 1.17 & 869.62 / 857.16 & 2227.72 / 1375.00 & 1.01x & 1.62x \\
Prefill & 256 & 1024 & 0.48 / 0.48 & 2090.36 / 2077.89 & 4832.36 / 2994.89 & 1.01x & 1.61x \\
Prefill & 1024 & 256 & 0.41 / 0.41 & 2444.41 / 2431.95 & 5570.82 / 3733.36 & 1.01x & 1.49x \\
Prefill & 1024 & 1024 & 0.26 / 0.26 & 3798.92 / 3786.46 & 8333.40 / 5437.21 & 1.00x & 1.53x \\
Decode & 256 & 256 & 7.12 / 14.21 & 140.46 / 70.35 & 556.20 / 278.22 & 2.00x & 2.00x \\
Decode & 256 & 1024 & 6.94 / 13.86 & 144.00 / 72.13 & 570.20 / 285.23 & 2.00x & 2.00x \\
Decode & 1024 & 256 & 6.94 / 13.86 & 144.00 / 72.13 & 570.20 / 285.23 & 2.00x & 2.00x \\
Decode & 1024 & 1024 & 6.78 / 13.53 & 147.53 / 73.90 & 584.20 / 292.23 & 2.00x & 2.00x \\
\bottomrule
\end{tabular}

%% file: tables/cr/w8a8_qwen2.tex
\begin{tabular}{lrrrrrrr}
\toprule
Stage & V tokens & T tokens & Throughput & Latency & Energy & Speedup & Energy ratio \\
\midrule
Prefill & 256 & 256 & 1.14 / 1.16 & 877.02 / 863.03 & 2400.87 / 1519.00 & 1.02$\times$ & 1.58$\times$ \\
Prefill & 256 & 1024 & 0.51 / 0.51 & 1977.79 / 1968.44 & 4817.34 / 3018.88 & 1.00$\times$ & 1.60$\times$ \\
Prefill & 1024 & 256 & 0.39 / 0.39 & 2559.89 / 2550.54 & 6045.12 / 4246.66 & 1.00$\times$ & 1.42$\times$ \\
Prefill & 1024 & 1024 & 0.27 / 0.27 & 3766.04 / 3756.69 & 8654.54 / 5838.09 & 1.00$\times$ & 1.48$\times$ \\
Decode & 256 & 256 & 8.03 / 15.95 & 124.55 / 62.71 & 493.04 / 246.94 & 1.99$\times$ & 2.00$\times$ \\
Decode & 256 & 1024 & 7.96 / 15.69 & 125.59 / 63.75 & 496.69 / 249.27 & 1.97$\times$ & 1.99$\times$ \\
Decode & 1024 & 256 & 7.96 / 15.69 & 125.59 / 63.75 & 496.69 / 249.27 & 1.97$\times$ & 1.99$\times$ \\
Decode & 1024 & 1024 & 7.90 / 15.43 & 126.63 / 64.79 & 500.34 / 251.60 & 1.95$\times$ & 1.99$\times$ \\
\bottomrule
\end{tabular}

%% file: tables/cr/w8a8_qwen25.tex
\begin{tabular}{lrrrrrrr}
\toprule
Stage & V tokens & T tokens & Throughput & Latency & Energy & Speedup & Energy ratio \\
\midrule
Prefill & 256 & 256 & 0.60 / 0.63 & 1663.57 / 1582.40 & 4802.11 / 2690.07 & 1.05$\times$ & 1.79$\times$ \\
Prefill & 256 & 1024 & 0.25 / 0.26 & 3959.49 / 3872.80 & 10486.84 / 5856.24 & 1.02$\times$ & 1.79$\times$ \\
Prefill & 1024 & 256 & 0.23 / 0.23 & 4439.11 / 4352.42 & 11419.90 / 6789.30 & 1.02$\times$ & 1.68$\times$ \\
Prefill & 1024 & 1024 & 0.14 / 0.15 & 6972.05 / 6814.82 & 17529.33 / 10164.20 & 1.02$\times$ & 1.72$\times$ \\
Decode & 256 & 256 & 4.02 / 8.00 & 248.69 / 125.07 & 984.06 / 492.73 & 1.99$\times$ & 2.00$\times$ \\
Decode & 256 & 1024 & 3.99 / 7.88 & 250.47 / 126.85 & 989.19 / 496.16 & 1.97$\times$ & 1.99$\times$ \\
Decode & 1024 & 256 & 3.99 / 7.88 & 250.47 / 126.85 & 989.19 / 496.16 & 1.97$\times$ & 1.99$\times$ \\
Decode & 1024 & 1024 & 3.96 / 7.77 & 252.25 / 128.63 & 994.31 / 499.58 & 1.96$\times$ & 1.99$\times$ \\
\bottomrule
\end{tabular}

%% file: tables/cr/deployment_ranges.tex
\begin{tabular}{lllrr}
\toprule
Model & Stage & Precision & Latency speedup $\uparrow$ & Energy reduction $\uparrow$ \\
\midrule
Qwen3-VL-2B & Prefill & W8A8 & 1.00--1.01x & 1.49--1.62x \\
Qwen3-VL-2B & Prefill & W4A8 & 1.58--1.76x & 2.20--2.70x \\
Qwen3-VL-2B & Prefill & W4A4 & 2.53--3.41x & 3.13--4.65x \\
Qwen3-VL-2B & Decode & W8A8 & 2.00x & 2.00x \\
Qwen3-VL-2B & Decode & W4A8 & 3.74--3.91x & 3.76--3.93x \\
Qwen3-VL-2B & Decode & W4A4 & 3.74--3.91x & 3.76--3.93x \\
Qwen2-VL-2B & Prefill & W8A8 & 1.00--1.02$\times$ & 1.42--1.60$\times$ \\
Qwen2-VL-2B & Prefill & W4A8 & 1.48--1.73$\times$ & 1.94--2.55$\times$ \\
Qwen2-VL-2B & Prefill & W4A4 & 2.11--3.13$\times$ & 2.53--4.14$\times$ \\
Qwen2-VL-2B & Decode & W8A8 & 1.95--1.99$\times$ & 1.99--2.00$\times$ \\
Qwen2-VL-2B & Decode & W4A8 & 3.74--3.92$\times$ & 3.88--3.97$\times$ \\
Qwen2-VL-2B & Decode & W4A4 & 3.92--3.97$\times$ & 3.93--3.98$\times$ \\
Qwen2.5-VL-3B & Prefill & W8A8 & 1.02--1.05$\times$ & 1.68--1.79$\times$ \\
Qwen2.5-VL-3B & Prefill & W4A8 & 1.70--1.89$\times$ & 2.67--3.13$\times$ \\
Qwen2.5-VL-3B & Prefill & W4A4 & 2.83--3.64$\times$ & 4.09--5.64$\times$ \\
Qwen2.5-VL-3B & Decode & W8A8 & 1.96--1.99$\times$ & 1.99--2.00$\times$ \\
Qwen2.5-VL-3B & Decode & W4A8 & 3.78--3.93$\times$ & 3.91--3.97$\times$ \\
Qwen2.5-VL-3B & Decode & W4A4 & 3.97--3.98$\times$ & 3.96--3.99$\times$ \\
\bottomrule
\end{tabular}

%% file: main.bib
@article{bubeck2023gpt4,
  title={Sparks of artificial general intelligence: Early experiments with gpt-4},
  author={Bubeck, S{\'e}bastien and Chandrasekaran, Varun and Eldan, Ronen and Gehrke, Johannes and Horvitz, Eric and Kamar, Ece and Lee, Peter and Lee, Yin Tat and Li, Yuanzhi and Lundberg, Scott and others},
  journal={arXiv preprint arXiv:2303.12712},
  year={2023}
}

@misc{huang2026modesacceleratingmixtureofexpertsmultimodal,
      title={MoDES: Accelerating Mixture-of-Experts Multimodal Large Language Models via Dynamic Expert Skipping}, 
      author={Yushi Huang and Zining Wang and Zhihang Yuan and Yifu Ding and Ruihao Gong and Jinyang Guo and Xianglong Liu and Jun Zhang},
      year={2026},
      eprint={2511.15690},
      archivePrefix={arXiv},
      primaryClass={cs.CV},
      url={https://arxiv.org/abs/2511.15690}, 
}

@article{gong2025pushing,
  title={Pushing the limit of post-training quantization},
  author={Gong, Ruihao and Liu, Xianglong and Li, Yuhang and Fan, Yunqiang and Wei, Xiuying and Guo, Jinyang},
  journal={IEEE Transactions on Pattern Analysis and Machine Intelligence},
  year={2025},
  publisher={IEEE}
}

@misc{lv2025llmcbenchmarkingvisionlanguagemodel,
      title={LLMC+: Benchmarking Vision-Language Model Compression with a Plug-and-play Toolkit}, 
      author={Chengtao Lv and Bilang Zhang and Yang Yong and Ruihao Gong and Yushi Huang and Shiqiao Gu and Jiajun Wu and Yumeng Shi and Jinyang Guo and Wenya Wang},
      year={2025},
      eprint={2508.09981},
      archivePrefix={arXiv},
      primaryClass={cs.CV},
      url={https://arxiv.org/abs/2508.09981}, 
}

@article{bai2025qwen25vl,
  title={Qwen2.5-VL Technical Report},
  author={Bai, Shuai and Chen, Keqin and Liu, Xuejing and Wang, Jialin and Ge, Wenbin and Song, Sibo and Dang, Kai and Wang, Peng and Wang, Shijie and Tang, Jun and Zhong, Humen and Zhu, Yuanzhi and Yang, Mingkun and Li, Zhaohai and Wan, Jianqiang and Wang, Pengfei and Ding, Wei and Fu, Zheren and Xu, Yiheng and Ye, Jiabo and Zhang, Xi and Xie, Tianbao and Cheng, Zesen and Zhang, Hang and Yang, Zhibo and Xu, Haiyang and Lin, Junyang},
  journal={arXiv preprint arXiv:2502.13923},
  year={2025}
}

@article{wang2024qwen2vl,
  title={Qwen2-VL: Enhancing Vision-Language Model's Perception of the World at Any Resolution},
  author={Wang, Peng and Bai, Shuai and Tan, Sinan and Wang, Shijie and Fan, Zhihao and Bai, Jinze and Chen, Keqin and Liu, Xuejing and Wang, Jialin and Ge, Wenbin and Fan, Yang and Dang, Kai and Du, Mengfei and Ren, Xuancheng and Men, Rui and Liu, Dayiheng and Zhou, Chang and Zhou, Jingren and Lin, Junyang},
  journal={arXiv preprint arXiv:2409.12191},
  year={2024}
}

@misc{xiaomi2025mimovl,
  title={MiMo-VL Technical Report}, 
  author={LLM-Core-Team Xiaomi},
  year={2025},
  eprint={2506.03569},
  archivePrefix={arXiv},
  primaryClass={cs.CL},
  url={https://arxiv.org/abs/2506.03569}, 
}

@inproceedings{liu2024llava1.5,
  title={Improved baselines with visual instruction tuning},
  author={Liu, Haotian and Li, Chunyuan and Li, Yuheng and Lee, Yong Jae},
  booktitle={Proceedings of the IEEE/CVF Conference on Computer Vision and Pattern Recognition},
  pages={26296--26306},
  year={2024}
}

@article{li2024llava-ov,
  title={Llava-onevision: Easy visual task transfer},
  author={Li, Bo and Zhang, Yuanhan and Guo, Dong and Zhang, Renrui and Li, Feng and Zhang, Hao and Zhang, Kaichen and Zhang, Peiyuan and Li, Yanwei and Liu, Ziwei and others},
  journal={arXiv preprint arXiv:2408.03326},
  year={2024}
}

@misc{zhu2025internvl3,
  title={InternVL3: Exploring Advanced Training and Test-Time Recipes for Open-Source Multimodal Models}, 
  author={Jinguo Zhu and Weiyun Wang and Zhe Chen and Zhaoyang Liu and Shenglong Ye and Lixin Gu and Hao Tian and Yuchen Duan and Weijie Su and Jie Shao and Zhangwei Gao and Erfei Cui and Xuehui Wang and Yue Cao and Yangzhou Liu and Xingguang Wei and Hongjie Zhang and Haomin Wang and Weiye Xu and Hao Li and Jiahao Wang and Nianchen Deng and Songze Li and Yinan He and Tan Jiang and Jiapeng Luo and Yi Wang and Conghui He and Botian Shi and Xingcheng Zhang and Wenqi Shao and Junjun He and Yingtong Xiong and Wenwen Qu and Peng Sun and Penglong Jiao and Han Lv and Lijun Wu and Kaipeng Zhang and Huipeng Deng and Jiaye Ge and Kai Chen and Limin Wang and Min Dou and Lewei Lu and Xizhou Zhu and Tong Lu and Dahua Lin and Yu Qiao and Jifeng Dai and Wenhai Wang},
  year={2025},
  eprint={2504.10479},
  archivePrefix={arXiv},
  primaryClass={cs.CV},
  url={https://arxiv.org/abs/2504.10479}, 
}

@article{bai2025qwen3vl,
    title={Qwen3-VL Technical Report}, 
    author={Shuai Bai and Yuxuan Cai and Ruizhe Chen and Keqin Chen and Xionghui Chen and Zesen Cheng and Lianghao Deng and Wei Ding and Chang Gao and Chunjiang Ge and Wenbin Ge and Zhifang Guo and Qidong Huang and Jie Huang and Fei Huang and Binyuan Hui and Shutong Jiang and Zhaohai Li and Mingsheng Li and Mei Li and Kaixin Li and Zicheng Lin and Junyang Lin and Xuejing Liu and Jiawei Liu and Chenglong Liu and Yang Liu and Dayiheng Liu and Shixuan Liu and Dunjie Lu and Ruilin Luo and Chenxu Lv and Rui Men and Lingchen Meng and Xuancheng Ren and Xingzhang Ren and Sibo Song and Yuchong Sun and Jun Tang and Jianhong Tu and Jianqiang Wan and Peng Wang and Pengfei Wang and Qiuyue Wang and Yuxuan Wang and Tianbao Xie and Yiheng Xu and Haiyang Xu and Jin Xu and Zhibo Yang and Mingkun Yang and Jianxin Yang and An Yang and Bowen Yu and Fei Zhang and Hang Zhang and Xi Zhang and Bo Zheng and Humen Zhong and Jingren Zhou and Fan Zhou and Jing Zhou and Yuanzhi Zhu and Ke Zhu},
    journal={arXiv preprint arXiv:2511.21631},
    year={2025}
}

@misc{zhang2024lmms-eval,
  title={LMMs-Eval: Reality Check on the Evaluation of Large Multimodal Models}, 
  author={Kaichen Zhang and Bo Li and Peiyuan Zhang and Fanyi Pu and Joshua Adrian Cahyono and Kairui Hu and Shuai Liu and Yuanhan Zhang and Jingkang Yang and Chunyuan Li and Ziwei Liu},
  year={2024},
  eprint={2407.12772},
  archivePrefix={arXiv},
  primaryClass={cs.CL},
  url={https://arxiv.org/abs/2407.12772}, 
}

@inproceedings{masry2022chartqa,
  title = "{C}hart{QA}: A Benchmark for Question Answering about Charts with Visual and Logical Reasoning",
  author = "Masry, Ahmed and Long, Do and Tan, Jia Qing and Joty, Shafiq and Hoque, Enamul",
  booktitle = "Findings of the Association for Computational Linguistics: ACL 2022",
  month = may,
  year = "2022",
  address = "Dublin, Ireland",
  publisher = "Association for Computational Linguistics",
  url = "https://aclanthology.org/2022.findings-acl.177",
  doi = "10.18653/v1/2022.findings-acl.177",
  pages = "2263--2279",
}

@inproceedings{mathew2021docvqa,
  title={Docvqa: A dataset for vqa on document images},
  author={Mathew, Minesh and Karatzas, Dimosthenis and Jawahar, CV},
  booktitle={Proceedings of the IEEE/CVF winter conference on applications of computer vision},
  pages={2200--2209},
  year={2021}
}

@inproceedings{lu2022scienceqa,
    title={Learn to Explain: Multimodal Reasoning via Thought Chains for Science Question Answering},
    author={Lu, Pan and Mishra, Swaroop and Xia, Tony and Qiu, Liang and Chang, Kai-Wei and Zhu, Song-Chun and Tafjord, Oyvind and Clark, Peter and Ashwin Kalyan},
    booktitle={The 36th Conference on Neural Information Processing Systems (NeurIPS)},
    year={2022}
}

@inproceedings{zellers2019hellaswag,
  title={Hellaswag: Can a machine really finish your sentence?},
  author={Zellers, Rowan and Holtzman, Ari and Bisk, Yonatan and Farhadi, Ali and Choi, Yejin},
  booktitle={Proceedings of the 57th annual meeting of the association for computational linguistics},
  pages={4791--4800},
  year={2019}
}

@misc{fu2023mme,
  title={MME: a comprehensive evaluation benchmark for multimodal large language models. CoRR abs/2306.13394 (2023)},
  author={Fu, Chaoyou and Chen, Peixian and Shen, Yunhang and Qin, Yulei and Zhang, Mengdan and Lin, Xu and Qiu, Zhenyu and Lin, Wei and Yang, Jinrui and Zheng, Xiawu and others},
  year={2023}
}

@inproceedings{kembhavi2016ai2d,
  title={A diagram is worth a dozen images},
  author={Kembhavi, Aniruddha and Salvato, Mike and Kolve, Eric and Seo, Minjoon and Hajishirzi, Hannaneh and Farhadi, Ali},
  booktitle={European conference on computer vision},
  pages={235--251},
  year={2016},
  organization={Springer}
}

@article{li2023seedbench,
  title={Seed-bench: Benchmarking multimodal llms with generative comprehension},
  author={Li, Bohao and Wang, Rui and Wang, Guangzhi and Ge, Yuying and Ge, Yixiao and Shan, Ying},
  journal={arXiv preprint arXiv:2307.16125},
  year={2023}
}

@inproceedings{yue2023mmmu,
    title={MMMU: A Massive Multi-discipline Multimodal Understanding and Reasoning Benchmark for Expert AGI},
    author={Xiang Yue and Yuansheng Ni and Kai Zhang and Tianyu Zheng and Ruoqi Liu and Ge Zhang and Samuel Stevens and Dongfu Jiang and Weiming Ren and Yuxuan Sun and Cong Wei and Botao Yu and Ruibin Yuan and Renliang Sun and Ming Yin and Boyuan Zheng and Zhenzhu Yang and Yibo Liu and Wenhao Huang and Huan Sun and Yu Su and Wenhu Chen},
    booktitle={Proceedings of CVPR},
    year={2024},
}

@misc{huang2025temporal-feature-matters,
      title={Temporal Feature Matters: A Framework for Diffusion Model Quantization}, 
      author={Yushi Huang and Ruihao Gong and Xianglong Liu and Jing Liu and Yuhang Li and Jiwen Lu and Dacheng Tao},
      year={2025},
      eprint={2407.19547},
      archivePrefix={arXiv},
      primaryClass={cs.CV},
      url={https://arxiv.org/abs/2407.19547}, 
}

@misc{huang2024tfmq-dm,
      title={TFMQ-DM: Temporal Feature Maintenance Quantization for Diffusion Models}, 
      author={Yushi Huang and Ruihao Gong and Jing Liu and Tianlong Chen and Xianglong Liu},
      year={2024},
      eprint={2311.16503},
      archivePrefix={arXiv},
      primaryClass={cs.CV},
      url={https://arxiv.org/abs/2311.16503}, 
}

@misc{huang2025qvgen,
      title={QVGen: Pushing the Limit of Quantized Video Generative Models}, 
      author={Yushi Huang and Ruihao Gong and Jing Liu and Yifu Ding and Chengtao Lv and Haotong Qin and Jun Zhang},
      year={2025},
      eprint={2505.11497},
      archivePrefix={arXiv},
      primaryClass={cs.CV},
      url={https://arxiv.org/abs/2505.11497}, 
}

@article{du2024bitdistiller,
  title={Bitdistiller: Unleashing the potential of sub-4-bit llms via self-distillation},
  author={Du, Dayou and Zhang, Yijia and Cao, Shijie and Guo, Jiaqi and Cao, Ting and Chu, Xiaowen and Xu, Ningyi},
  journal={arXiv preprint arXiv:2402.10631},
  year={2024}
}

@article{liu2023llm,
  title={Llm-qat: Data-free quantization aware training for large language models},
  author={Liu, Zechun and Oguz, Barlas and Zhao, Changsheng and Chang, Ernie and Stock, Pierre and Mehdad, Yashar and Shi, Yangyang and Krishnamoorthi, Raghuraman and Chandra, Vikas},
  journal={arXiv preprint arXiv:2305.17888},
  year={2023}
}

@article{li2021brecq,
  title={Brecq: Pushing the limit of post-training quantization by block reconstruction},
  author={Li, Yuhang and Gong, Ruihao and Tan, Xu and Yang, Yang and Hu, Peng and Zhang, Qi and Yu, Fengwei and Wang, Wei and Gu, Shi},
  journal={arXiv preprint arXiv:2102.05426},
  year={2021}
}

@inproceedings{nagel2020adaround,
  title={Up or down? adaptive rounding for post-training quantization},
  author={Nagel, Markus and Amjad, Rana Ali and Van Baalen, Mart and Louizos, Christos and Blankevoort, Tijmen},
  booktitle={International Conference on Machine Learning},
  pages={7197--7206},
  year={2020},
  organization={PMLR}
}

@inproceedings{xiao2023smoothquant,
  title={Smoothquant: Accurate and efficient post-training quantization for large language models},
  author={Xiao, Guangxuan and Lin, Ji and Seznec, Mickael and Wu, Hao and Demouth, Julien and Han, Song},
  booktitle={International Conference on Machine Learning},
  pages={38087--38099},
  year={2023},
  organization={PMLR}
}

@article{frantar2022gptq,
  title={Gptq: Accurate post-training quantization for generative pre-trained transformers},
  author={Frantar, Elias and Ashkboos, Saleh and Hoefler, Torsten and Alistarh, Dan},
  journal={arXiv preprint arXiv:2210.17323},
  year={2022}
}

@article{lin2023awq,
  title={AWQ: Activation-aware Weight Quantization for LLM Compression and Acceleration},
  author={Lin, Ji and Tang, Jiaming and Tang, Haotian and Yang, Shang and Dang, Xingyu and Han, Song},
  journal={arXiv preprint arXiv:2306.00978},
  year={2023}
}

@article{huang2024billm,
  title={Billm: Pushing the limit of post-training quantization for llms},
  author={Huang, Wei and Liu, Yangdong and Qin, Haotong and Li, Ying and Zhang, Shiming and Liu, Xianglong and Magno, Michele and Qi, Xiaojuan},
  journal={arXiv preprint arXiv:2402.04291},
  year={2024}
}

@article{shao2023omniquant,
  title={Omniquant: Omnidirectionally calibrated quantization for large language models},
  author={Shao, Wenqi and Chen, Mengzhao and Zhang, Zhaoyang and Xu, Peng and Zhao, Lirui and Li, Zhiqian and Zhang, Kaipeng and Gao, Peng and Qiao, Yu and Luo, Ping},
  journal={arXiv preprint arXiv:2308.13137},
  year={2023}
}

@article{li2025gptqv2,
  title={GPTAQ: Efficient Finetuning-Free Quantization for Asymmetric Calibration}, 
  author={Yuhang Li and Ruokai Yin and Donghyun Lee and Shiting Xiao and Priyadarshini Panda},
  year={2025},
  journal={arXiv preprint arXiv:2504.02692},
}

@misc{gong2024llmc,
  title={LLMC: Benchmarking Large Language Model Quantization with a Versatile Compression Toolkit},
  author={Ruihao Gong and Yang Yong and Shiqiao Gu and Yushi Huang and Chentao Lv and Yunchen Zhang and Xianglong Liu and Dacheng Tao},
  year={2024},
  eprint={2405.06001},
  archivePrefix={arXiv},
  primaryClass={cs.LG},
  url={https://arxiv.org/abs/2405.06001},
}

@article{ashkboos2024quarot,
  title={QuaRot: Outlier-Free 4-Bit Inference in Rotated LLMs},
  author={Ashkboos, Saleh and Mohtashami, Amirkeivan and Croci, Maximilian L and Li, Bo and Jaggi, Martin and Alistarh, Dan and Hoefler, Torsten and Hensman, James},
  journal={arXiv preprint arXiv:2404.00456},
  year={2024}
}

@article{liu2024spinquant,
  title={Spinquant: Llm quantization with learned rotations},
  author={Liu, Zechun and Zhao, Changsheng and Fedorov, Igor and Soran, Bilge and Choudhary, Dhruv and Krishnamoorthi, Raghuraman and Chandra, Vikas and Tian, Yuandong and Blankevoort, Tijmen},
  journal={arXiv preprint arXiv:2405.16406},
  year={2024}
}

@article{hu2025ostquant,
  title={Ost{Q}uant: Refining Large Language Model Quantization with Orthogonal and Scaling Transformations for Better Distribution Fitting},
  author={Hu, Xing and Cheng, Yuan and Yang, Dawei and Xu, Zukang and Yuan, Zhihang and Yu, Jiangyong and Xu, Chen and Jiang, Zhe and Zhou, Sifan},
  journal={arXiv preprint arXiv:2501.13987},
  year={2025}
}

@article{sun2024flatquant,
  title={Flatquant: Flatness matters for llm quantization},
  author={Sun, Yuxuan and Liu, Ruikang and Bai, Haoli and Bao, Han and Zhao, Kang and Li, Yuening and Hu, Jiaxin and Yu, Xianzhi and Hou, Lu and Yuan, Chun and others},
  journal={arXiv preprint arXiv:2410.09426},
  year={2024}
}

@article{yu2025mquant,
  title={MQuant: Unleashing the Inference Potential of Multimodal Large Language Models via Full Static Quantization},
  author={Yu, JiangYong and Zhou, Sifan and Yang, Dawei and Wang, Shuo and Li, Shuoyu and Hu, Xing and Xu, Chen and Xu, Zukang and Shu, Changyong and Yuan, Zhihang},
  journal={arXiv preprint arXiv:2502.00425},
  year={2025}
}

@inproceedings{li2025mbq,
  title={Mbq: Modality-balanced quantization for large vision-language models},
  author={Li, Shiyao and Hu, Yingchun and Ning, Xuefei and Liu, Xihui and Hong, Ke and Jia, Xiaotao and Li, Xiuhong and Yan, Yaqi and Ran, Pei and Dai, Guohao and others},
  booktitle={Proceedings of the Computer Vision and Pattern Recognition Conference},
  pages={4167--4177},
  year={2025}
}

@article{wang2024q-vlm,
  title={Q-VLM: Post-training Quantization for Large Vision-Language Models},
  author={Wang, Changyuan and Wang, Ziwei and Xu, Xiuwei and Tang, Yansong and Zhou, Jie and Lu, Jiwen},
  journal={arXiv preprint arXiv:2410.08119},
  year={2024}
}

@inproceedings{xie2024qslaw,
  title={Advancing Multimodal Large Language Models with Quantization-Aware Scale Learning for Efficient Adaptation},
  author={Xie, Jingjing and Zhang, Yuxin and Lin, Mingbao and Cao, Liujuan and Ji, Rongrong},
  booktitle={Proceedings of the 32nd ACM International Conference on Multimedia},
  pages={10582--10591},
  year={2024}
}

@misc{wu2024deepseekvl2,
      title={DeepSeek-VL2: Mixture-of-Experts Vision-Language Models for Advanced Multimodal Understanding}, 
      author={Zhiyu Wu and Xiaokang Chen and Zizheng Pan and Xingchao Liu and Wen Liu and Damai Dai and Huazuo Gao and Yiyang Ma and Chengyue Wu and Bingxuan Wang and Zhenda Xie and Yu Wu and Kai Hu and Jiawei Wang and Yaofeng Sun and Yukun Li and Yishi Piao and Kang Guan and Aixin Liu and Xin Xie and Yuxiang You and Kai Dong and Xingkai Yu and Haowei Zhang and Liang Zhao and Yisong Wang and Chong Ruan},
      year={2024},
      eprint={2412.10302},
      archivePrefix={arXiv},
      primaryClass={cs.CV},
      url={https://arxiv.org/abs/2412.10302}, 
}

@article{chen2025qat-scaling-law,
  title={Scaling law for quantization-aware training},
  author={Chen, Mengzhao and Zhang, Chaoyi and Liu, Jing and Zeng, Yutao and Xue, Zeyue and Liu, Zhiheng and Li, Yunshui and Ma, Jin and Huang, Jie and Zhou, Xun and others},
  journal={arXiv preprint arXiv:2505.14302},
  year={2025}
}

@article{xue2025vlmq,
  title={Vlmq: Efficient post-training quantization for large vision-language models via hessian augmentation},
  author={Xue, Yufei and Huang, Yushi and Shao, Jiawei and Zhang, Jun},
  journal={arXiv preprint arXiv:2508.03351},
  year={2025}
}

@inproceedings{chen2025efficientqat,
  title={Efficientqat: Efficient quantization-aware training for large language models},
  author={Chen, Mengzhao and Shao, Wenqi and Xu, Peng and Wang, Jiahao and Gao, Peng and Zhang, Kaipeng and Luo, Ping},
  booktitle={Proceedings of the 63rd Annual Meeting of the Association for Computational Linguistics (Volume 1: Long Papers)},
  pages={10081--10100},
  year={2025}
}

@article{guo2025speedq,
  title={SPEED-Q: Staged Processing with Enhanced Distillation towards Efficient Low-bit On-device VLM Quantization},
  author={Guo, Tianyu and Zhao, Shanwei and Zhu, Shiai and Ma, Chenguang},
  journal={arXiv preprint arXiv:2511.08914},
  year={2025}
}

@article{tian2024drivevlm,
  title={Drivevlm: The convergence of autonomous driving and large vision-language models},
  author={Tian, Xiaoyu and Gu, Junru and Li, Bailin and Liu, Yicheng and Wang, Yang and Zhao, Zhiyong and Zhan, Kun and Jia, Peng and Lang, Xianpeng and Zhao, Hang},
  journal={arXiv preprint arXiv:2402.12289},
  year={2024}
}

@inproceedings{chen2024videollm-online,
  title={Videollm-online: Online video large language model for streaming video},
  author={Chen, Joya and Lv, Zhaoyang and Wu, Shiwei and Lin, Kevin Qinghong and Song, Chenan and Gao, Difei and Liu, Jia-Wei and Gao, Ziteng and Mao, Dongxing and Shou, Mike Zheng},
  booktitle={Proceedings of the IEEE/CVF Conference on Computer Vision and Pattern Recognition},
  pages={18407--18418},
  year={2024}
}

@incollection{gholami2022quant-survey,
  title={A survey of quantization methods for efficient neural network inference},
  author={Gholami, Amir and Kim, Sehoon and Dong, Zhen and Yao, Zhewei and Mahoney, Michael W and Keutzer, Kurt},
  booktitle={Low-power computer vision},
  pages={291--326},
  year={2022},
  publisher={Chapman and Hall/CRC}
}

@article{bengio2013ste,
  title={Estimating or propagating gradients through stochastic neurons for conditional computation},
  author={Bengio, Yoshua and L{\'e}onard, Nicholas and Courville, Aaron},
  journal={arXiv preprint arXiv:1308.3432},
  year={2013}
}

@article{li2024llm-hardware-accelerator,
  title={Large language model inference acceleration: A comprehensive hardware perspective},
  author={Li, Jinhao and Xu, Jiaming and Huang, Shan and Chen, Yonghua and Li, Wen and Liu, Jun and Lian, Yaoxiu and Pan, Jiayi and Ding, Li and Zhou, Hao and others},
  journal={arXiv preprint arXiv:2410.04466},
  year={2024}
}

@article{van2023fp8-vs-int8,
  title={FP8 versus INT8 for efficient deep learning inference},
  author={Van Baalen, Mart and Kuzmin, Andrey and Nair, Suparna S and Ren, Yuwei and Mahurin, Eric and Patel, Chirag and Subramanian, Sundar and Lee, Sanghyuk and Nagel, Markus and Soriaga, Joseph and others},
  journal={arXiv preprint arXiv:2303.17951},
  year={2023}
}

@article{shao2025aiflow,
  title={Ai flow at the network edge},
  author={Shao, Jiawei and Li, Xuelong},
  journal={IEEE Network},
  year={2025},
  publisher={IEEE}
}

@article{an2026aiflow,
  title={Ai flow: Perspectives, scenarios, and approaches},
  author={An, Hongjun and Hu, Wenhan and Huang, Sida and Huang, Siqi and Li, Ruanjun and Liang, Yuanzhi and Shao, Jiawei and Song, Yiliang and Wang, Zihan and Yuan, Cheng and others},
  journal={Vicinagearth},
  volume={3},
  number={1},
  pages={1},
  year={2026},
  publisher={Springer}
}

@inproceedings{kwon2023vllm,
  title={Efficient Memory Management for Large Language Model Serving with PagedAttention},
  author={Woosuk Kwon and Zhuohan Li and Siyuan Zhuang and Ying Sheng and Lianmin Zheng and Cody Hao Yu and Joseph E. Gonzalez and Hao Zhang and Ion Stoica},
  booktitle={Proceedings of the ACM SIGOPS 29th Symposium on Operating Systems Principles},
  year={2023}
}

@inproceedings{dong2026isscc,
  author={Dong, Pingcheng and Tan, Yonghao and Liu, Xuejiao and Luo, Peng and Liu, Yu and Pang, Di and Ma, Songchen and Huang, Xijie and Liu, Shih-Yang and Zhang, Dong and Lu, Zhichao and Liang, Luhong and Tsui, Chi-Ying and Tu, Fengbin and Zhao, Liang and Cheng, Kwang-Ting},
  booktitle={2026 IEEE International Solid-State Circuits Conference (ISSCC)}, 
  title={31.1 A 14.08-to-135.69Token/s ReRAM-on-Logic Stacked Outlier-Free Large-Language-Model Accelerator with Block-Clustered Weight-Compression and Adaptive Parallel-Speculative-Decoding}, 
  year={2026},
  volume={69},
  number={},
  pages={532-534},
  doi={10.1109/ISSCC49663.2026.11409211}
}

@inproceedings{kim2026isscc,
  author={Kim, Sangjin and Oh, Jungjun and Kim, Byeongcheol and Choi, Yuseon and Park, Gwangtae and Yoo, Hoi-Jun},
  booktitle={2026 IEEE International Solid-State Circuits Conference (ISSCC)}, 
  title={31.2 Revolver: Low-Bit GenAI Accelerator for Distilled-Model and CoT with Phase-Aware-Quantization and Rotation-Based Integer-Scaled Group Quantization}, 
  year={2026},
  volume={69},
  number={},
  pages={534-536},
  doi={10.1109/ISSCC49663.2026.1140901}
}

@inproceedings{liu2026isscc,
  author={Liu, Bo and Zou, Zihan and Yan, Xinming and Kang, Xilong and Chen, Xinyang and Hu, Bo and Lin, Jiaming and Du, Haoran and Yang, Jun and Si, Xin and Cai, Hao},
  booktitle={2026 IEEE International Solid-State Circuits Conference (ISSCC)}, 
  title={A 51.6uJ/Token Subspace-Rotation-Based Dual-Quantized Large-language-Model Accelerator with Fused Scale-Activation INT Datapath and Rearranged Bit-Slice LUT Computation}, 
  year={2026},
  volume={69},
  number={},
  pages={536-538},
  doi={10.1109/ISSCC49663.2026.11409212}
}

@article{yuan2025task,
  title={Task-oriented feature compression for multimodal understanding via device-edge co-inference},
  author={Yuan, Cheng and Liu, Zhening and Lv, Jiashu and Shao, Jiawei and Jiang, Yufei and Zhang, Jun and Li, Xuelong},
  journal={IEEE Transactions on Mobile Computing},
  year={2025},
  publisher={IEEE}
}

@article{li2026token,
  title={Token-level Data Selection for Safe LLM Fine-tuning},
  author={Li, Yanping and Liu, Zhening and Li, Zijian and Lin, Zehong and Zhang, Jun},
  journal={arXiv preprint arXiv:2603.01185},
  year={2026}
}

@inproceedings{vlmsarebiased,
  title     = {Vision Language Models are Biased},
  author    = {An Vo and Khai-Nguyen Nguyen and Mohammad Reza Taesiri and Vy Tuong Dang and Anh Totti Nguyen and Daeyoung Kim},
  booktitle = {The Fourteenth International Conference on Learning Representations (ICLR)},
  year      = {2026},
  url       = {https://arxiv.org/abs/2505.23941}
}

@inproceedings{guan2024hallusionbench,
  title={Hallusionbench: an advanced diagnostic suite for entangled language hallucination and visual illusion in large vision-language models},
  author={Guan, Tianrui and Liu, Fuxiao and Wu, Xiyang and Xian, Ruiqi and Li, Zongxia and Liu, Xiaoyu and Wang, Xijun and Chen, Lichang and Huang, Furong and Yacoob, Yaser and others},
  booktitle={2024 IEEE/CVF Conference on Computer Vision and Pattern Recognition (CVPR)},
  pages={14375--14385},
  year={2024},
  organization={IEEE}
}

@article{li2023towards,
  title={Towards general text embeddings with multi-stage contrastive learning},
  author={Li, Zehan and Zhang, Xin and Zhang, Yanzhao and Long, Dingkun and Xie, Pengjun and Zhang, Meishan},
  journal={arXiv preprint arXiv:2308.03281},
  year={2023}
}

@article{team2026ovis,
  title={Ovis-Embedding: Pushing the Frontiers of Universal Omni-Modal Embeddings},
  author={Team, Ovis-Embedding},
  journal={arXiv preprint arXiv:2609.25165},
  year={2026}
}

@article{li2026pace,
  title={PACE: Progressive Angular-to-Norm Contrastive Embedding},
  author={Li, Yanping and Zhou, Wei and Liu, Yawen and Wang, Yibo and Zhu, Ke and Huzhang, Guangda and Chen, Qing-Guo and Xu, Zhao and Zhang, Jun and Wei, Wei},
  journal={arXiv preprint arXiv:2609.15152},
  year={2026}
}

@article{li2026gear,
  title={GEAR: Granularity-Adaptive Advantage Reweighting for LLM Agents via Self-Distillation},
  author={Li, Sijia and Huang, Yuchen and Liu, Zifan and Li, Yanping and Fu, Jingjing and Zhao, Li and Bian, Jiang and Zhang, Ling and Zhang, Jun and Wang, Rui},
  journal={arXiv preprint arXiv:2605.11853},
  year={2026}
}
